\documentclass[10pt,twocolumn,letterpaper]{article}

\usepackage{cvpr}              

\usepackage[dvipsnames]{xcolor}

\usepackage{multirow}
\usepackage{colortbl}  
\usepackage{xcolor}
\usepackage{array}   
\usepackage{booktabs}
\usepackage{indentfirst}
\usepackage{graphicx}
\usepackage{blindtext}
\usepackage{caption}
\usepackage{subcaption}
\usepackage{ntheorem}
\newtheorem*{thry_algorithm}{}
\usepackage[ruled,vlined]{algorithm2e}
\usepackage{stfloats}
\usepackage[accsupp]{axessibility}

\definecolor{cvprblue}{rgb}{0.21,0.49,0.74}
\usepackage[pagebackref,breaklinks,colorlinks,citecolor=cvprblue]{hyperref}

\def\paperID{3150} 
\def\confName{CVPR}
\def\confYear{2024}

\title{Stronger, Fewer, \& Superior: Harnessing Vision Foundation Models \\for Domain Generalized Semantic Segmentation}
\author{%
Zhixiang Wei{$^{1}$\thanks{Equal contribution.}} ~ Lin Chen{$^{1,2*}$}~ Yi Jin{$^{1*}$} ~ Xiaoxiao Ma{$^{1}$} ~ Tianle Liu{$^{1}$}~ Pengyang Ling{$^{1,2}$}~ Ben Wang{$^{1}$}~ \\
Huaian Chen{$^{1}$\footnotemark[2]} ~ Jinjin Zheng{$^{1}$}\\
\normalsize
$^{1}$\	University of Science and Technology of China ~~ $^{2}$\,Shanghai AI Laboratory \\
\normalsize
{\tt\small \{zhixiangwei,chlin,xiao\_xiao,tleliu,lpyang27,wblzgrsn,anchen\}@mail.ustc.edu.cn} \\
{\tt\small \{jinyi08,jjzheng\}@ustc.edu.cn}
}

\begin{document}
\twocolumn[{%
	\renewcommand\twocolumn[1][]{#1}%
	\maketitle%
    \setlength{\abovecaptionskip}{0.1cm}
    \setlength{\belowcaptionskip}{0.1cm}
	\begin{center}
		\centering
        \vspace{-0.6cm}
        \begin{thry_algorithm}\label{fig:figure1}
        \end{thry_algorithm}
    \begin{tabular}{c@{\extracolsep{0.0em}}c@{\extracolsep{0.7em}}c} 
		\includegraphics[width=0.2255\textwidth]{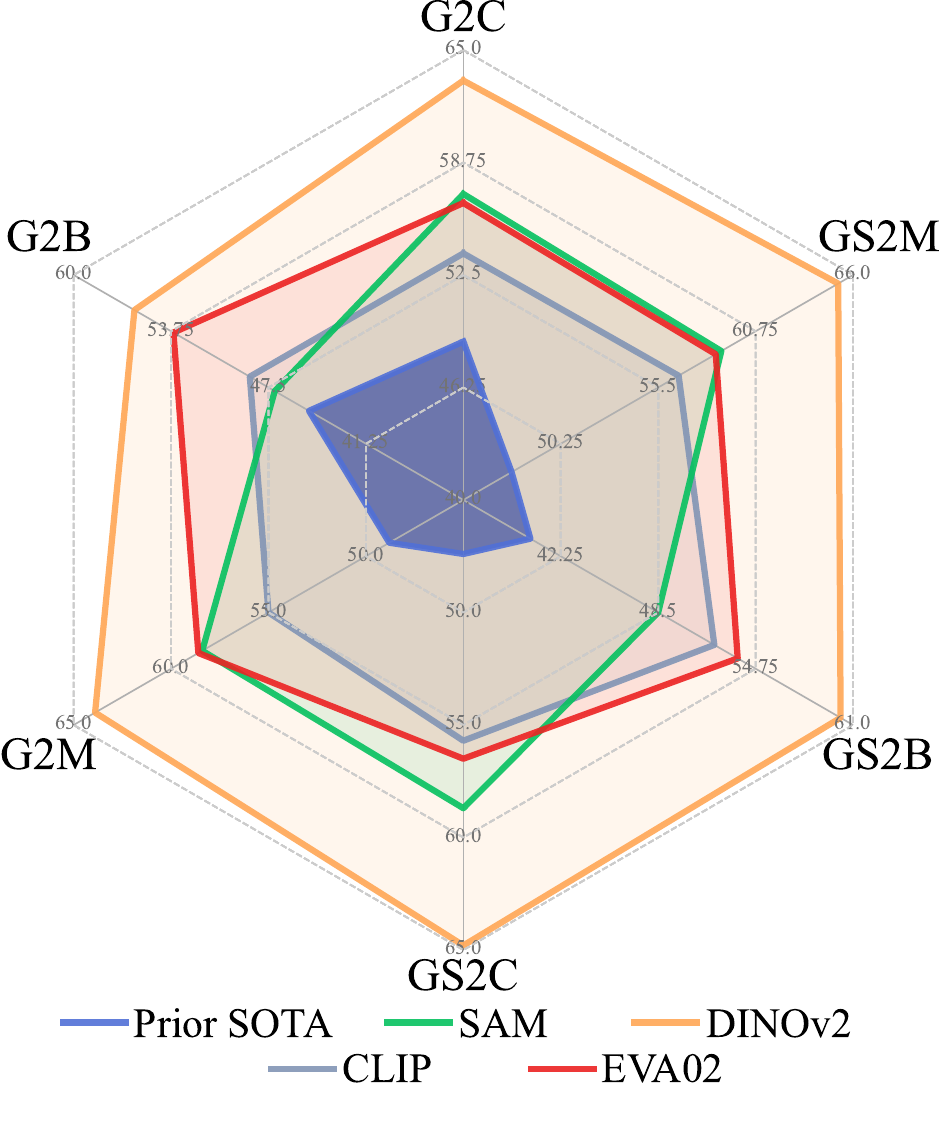}&
		\includegraphics[width=0.315\textwidth]{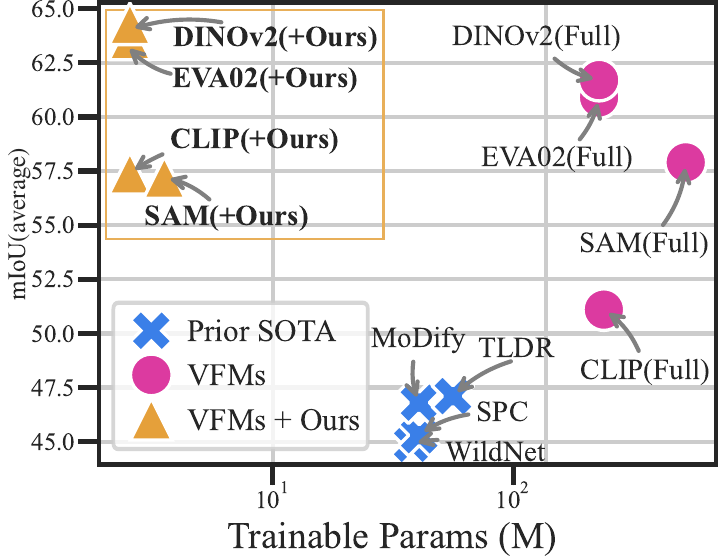}&
		\includegraphics[width=0.46\textwidth]{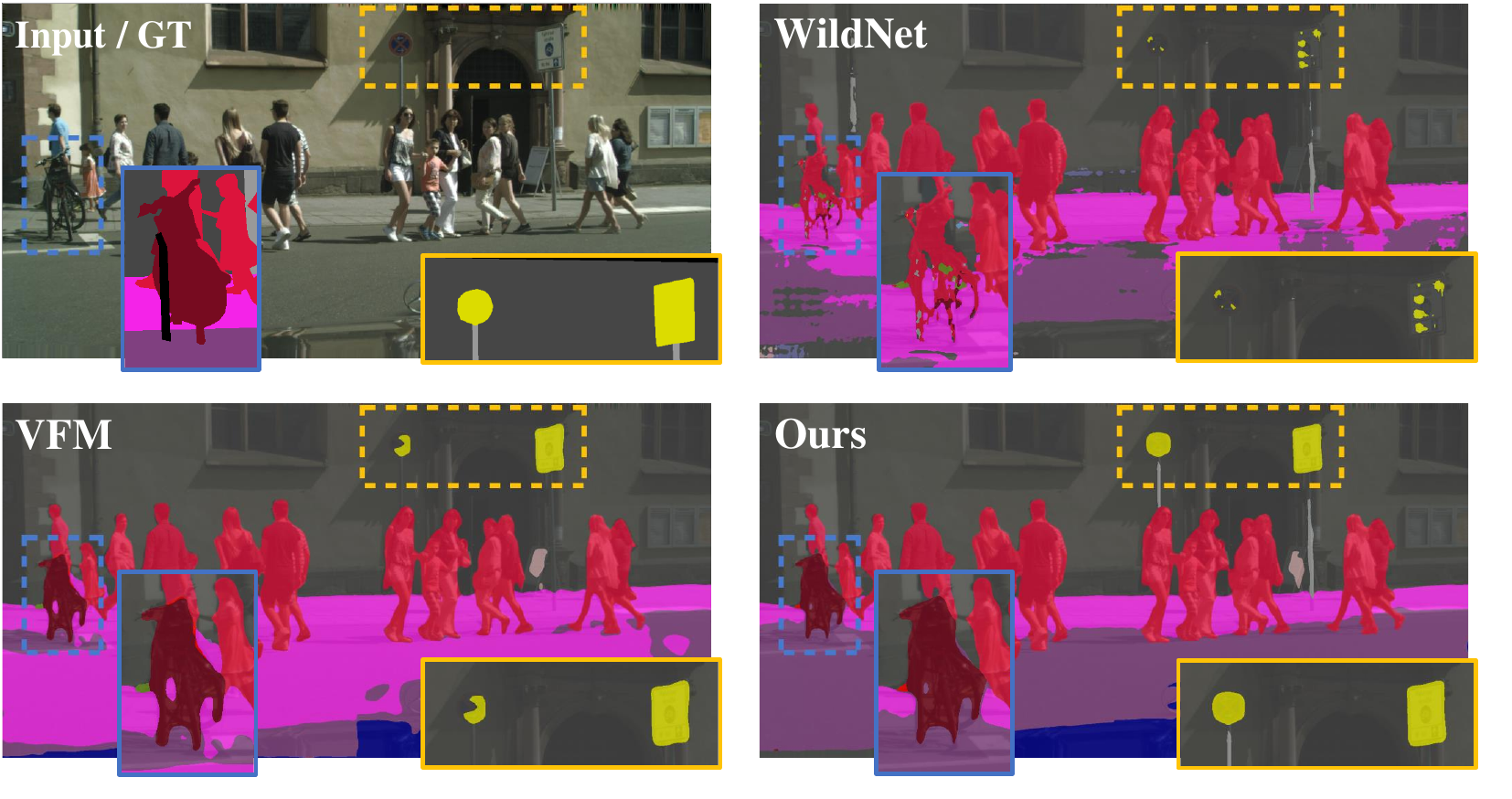}\\
		\footnotesize (a) \textbf{Stronger} pre-trained models&	\footnotesize ~~~~~~~~(b) \textbf{Fewer} trainable parameters&	\footnotesize (c) \textbf{Superior} generalization ability\\
	\end{tabular}
    \captionof{figure}{Vision Foundation Models (VFMs) are \textbf{stronger} pre-trained models that serve as robust backbones, effortlessly outperforming previous state-of-the-art Domain Generalized Semantic Segmentation (DGSS), as shown in (a). Yet, the extensive parameters of VFMs make them challenging to train. To address this, we introduce a robust fine-tuning approach to efficiently harness VFMs for DGSS. As illustrated in (b) and (c), the proposed methods achieve \textbf{superior} generalizability with \textbf{fewer} trainable parameters within backbones.}
    \end{center}     
}]
\renewcommand{\thefootnote}{\fnsymbol{footnote}}
\footnotetext[1]{~indicates equal contributions.}
\footnotetext[2]{~Corresponding authors.}
\begin{abstract}
\vspace{-1mm}
In this paper, we first assess and harness various Vision Foundation Models (VFMs) in the context of Domain Generalized Semantic Segmentation (DGSS). 
Driven by the motivation that \textbf{Leveraging Stronger pre-trained models and Fewer trainable parameters for Superior generalizability}, we introduce a robust fine-tuning approach, namely ``\textbf{Rein}", to parameter-efficiently harness VFMs for DGSS. 
Built upon a set of trainable tokens, each linked to distinct instances, Rein precisely refines and forwards the feature maps from each layer to the next layer within the backbone. This process produces diverse refinements for different categories within a single image. With fewer trainable parameters, Rein efficiently fine-tunes VFMs for DGSS tasks, surprisingly surpassing full parameter fine-tuning.
Extensive experiments across various settings demonstrate that Rein significantly outperforms state-of-the-art methods. Remarkably, with just an extra \textbf{1\%} of trainable parameters within the frozen backbone, Rein achieves a mIoU of \textbf{78.4\%} on the Cityscapes, without accessing any real urban-scene datasets. 
Code is available at \url{https://github.com/w1oves/Rein.git}. 
\vspace{5.5mm}
\end{abstract}
\begin{table*}[htbp]
    \centering
    \setlength{\abovecaptionskip}{0.1cm}
    \setlength{\belowcaptionskip}{0.1cm}
    \resizebox{\textwidth}{!}{
    \renewcommand\arraystretch{1.1}
    \setlength\tabcolsep{2.0pt}{
        \begin{tabular}{l|cccccc|ccccc}
            \hline
                         & \multicolumn{6}{c}{Previous DGSS methods} & \multicolumn{5}{|c}{Frozen backbone of VFMs}                                                                                                                                                                                                                        \\
            \hline
            Methods      &GTR\cite{gtrltr}&AdvStyle\cite{advstyle}& WildNet\cite{wildnet}                     & SPC\cite{SPC}                                 & PASTA\cite{PASTA} & TLDR\cite{TLDR} & CLIP-ViT-L\cite{CLIP}        & MAE-L\cite{MAE}              & SAM-H\cite{SAM}              & EVA02-L\cite{EVA02}            & DINOv2-L\cite{Dinov2}        \\
            Publications &TIP21&NIPS22& CVPR22                                    & CVPR23                                                    & ICCV23            & ICCV23         & ICML21                       & CVPR22                       & ICCV23                       & arXiv23                       & arXiv23                      \\
            \hline
            mIoU (Citys) &43.7&43.4& 45.8                                      & 46.7                                                      & 45.3              & 47.6            & 53.7\cellcolor[HTML]{EFEFEF} & 43.3\cellcolor[HTML]{EFEFEF} & 57.0\cellcolor[HTML]{EFEFEF} & 56.5\cellcolor[HTML]{EFEFEF} & \textbf{63.3}\cellcolor[HTML]{EFEFEF} \\
            mIoU (BDD)   &39.6&40.3& 41.7                                      & 43.7                                                       & 42.3              & 44.9            & 48.7\cellcolor[HTML]{EFEFEF} & 37.8\cellcolor[HTML]{EFEFEF} & 47.1\cellcolor[HTML]{EFEFEF} & 53.6\cellcolor[HTML]{EFEFEF} & \textbf{56.1}\cellcolor[HTML]{EFEFEF} \\
            mIoU (Map)   &39.1&42.0& 47.1                                      & 45.5                                                      & 48.6              & 48.8            & 55.0\cellcolor[HTML]{EFEFEF} & 48.0\cellcolor[HTML]{EFEFEF} & 58.4\cellcolor[HTML]{EFEFEF} & 58.6\cellcolor[HTML]{EFEFEF} & \textbf{63.9}\cellcolor[HTML]{EFEFEF} \\
            \hline
            mIoU (Average)  &40.8&41.9& 44.9                                      & 45.3                                                      & 45.4              & 47.1            & 52.4\cellcolor[HTML]{EFEFEF} & 43.0\cellcolor[HTML]{EFEFEF} & 54.2\cellcolor[HTML]{EFEFEF} & 56.2\cellcolor[HTML]{EFEFEF} & \textbf{61.1}\cellcolor[HTML]{EFEFEF} \\
            \hline
        \end{tabular}
    }}
    \caption{Performance benchmarking of \textbf{multiple VFMs and previous DGSS methods} under the \textit{GTAV $\rightarrow$ Cityscapes (Citys) + BDD100K (BDD) + Mapillary (Map)} generalization setting. Without specialized design, frozen VFMs demonstrate \textbf{stronger} performance.}
    \label{tab:inversitigate}
    \vspace{-4mm}
\end{table*}
\vspace{-13.0 mm}
\section{Introduction}
\label{sec:intro}
Prior works~\cite{dg:kang2022style,dg:kim2023single,dg:kim2023wedge,dg:reddy2022master,dg:termohlen2023re,dg:wu2022siamdoge,dg:zhang2023learning} in Domain Generalized Semantic Segmentation (DGSS) focus on improving prediction accuracy across multiple unseen domains without accessing their data, thus enabling a high generalization for real applications.
Since models are fine-tuned using datasets~\cite{cityscapes,gtav} that are either limited in scale or different in image style from the target domain, complex data augmentation approaches~\cite{advstyle,PASTA,gtrltr} and domain invariant feature extraction strategies~\cite{dg:xu2022dirl,ibn,choi2021robustnet,dg:tang2020selfnorm} have been widely explored in previous DGSS. These methods result in enhanced generalization when applied to classic backbones, \eg, VGGNet~\cite{vggnet}, MobileNetV2~\cite{mobilenet}, and ResNet~\cite{resnet}.

In recent years, large-scale Vision Foundation Models (VFMs) like CLIP~\cite{CLIP}, MAE~\cite{MAE}, SAM~\cite{SAM}, EVA02~\cite{EVA,EVA02}, and DINOv2~\cite{Dinov2} have significantly advanced the boundaries of performance in a variety of computer vision challenges. Giving the remarkable generalization of these VFMs across various unseen scenes, two intuitive questions emerge: \textit{How do VFMs perform in the context of DGSS?} And \textit{How to harness VFMs for DGSS?} We attempt to answer these questions as follows:

\textbf{Stronger}: We begin by evaluating and comparing the performance of various VFMs against existing DGSS methods. To ensure a fair comparison, we use image encoders from a variety of VFMs as the backbone for feature extraction in all cases. These backbones are coupled with the widely-used decode head, \ie, Mask2Former~\cite{mask2former}, to generate semantic predictions. As illustrated in Tab.~\ref{tab:inversitigate}, while previous DGSS methods have showcased commendable results, they perform less effectively compared to frozen VFMs. This finding clearly demonstrates the powerful potential of VFMs in DGSS, outperforming traditional backbones like ResNet~\cite{resnet} and MobileNetV2~\cite{mobilenet}, thereby establishing VFMs as a meaningful benchmark in the field.

\textbf{Fewer}: Although VFMs have exhibited impressive generalization capabilities, fine-tuning them for DGSS tasks poses a challenge. The datasets~\cite{gtav,cityscapes} commonly used in DGSS tasks are significantly smaller in scale compared to ImageNet~\cite{imagenet}, and fine-tuning VFMs with their huge number of trainable parameters on these datasets result in limited generalizability~\cite{scaling}. To address this issue, instead of the difficult task of large datasets collection, we resort to fine-tuning VFMs with fewer trainable parameters. However, most existing parameter-efficient fine-tuning strategies, which fine-tune a large-scale model with fewer trainable parameters, are primarily designed for adapting large language models~\cite{lora,Bitfit,adapter,prompt_tuning,peft:prefix,peft:p_tuning,peft:prompt_for_vision_language_models} or classification networks~\cite{adaptformer,vpt}. These methods are not developed for refining features for distinct instances within a single image, thereby limiting their effectiveness in DGSS contexts.

\textbf{Superior}: In this work, we introduce a robust and efficient fine-tuning approach, namely ``Rein". Tailored for DGSS tasks, Rein employs fewer trainable parameters to harness stronger VFMs for achieving superior generalization. At its core, Rein comprises a set of randomly initialized tokens, each directly linked to different instances. These tokens, through a dot-product operation with VFMs features, generate an attention-like similarity map. This map enables Rein to perform precise refinement tailored to each instance within an image, significantly boosting VFMs in the context of DGSS. Moreover, to reduce the number of trainable parameters, we employ shared weights across MLPs in different layers and design our learnable tokens by multiplying two low-rank matrices. Extensive experiments on various DGSS settings demonstrate that the proposed Rein outperforms existing DGSS methods by a large margin with fewer trainable parameters. In a nutshell, the \textbf{main contributions} of this paper are as follows:
\begin{itemize}
    \item We first assess various Vision Foundation Models (VFMs) in the context of Domain Generalized Semantic Segmentation (DGSS). Our extensive experiments in the DGSS framework highlight the impressive generalization capabilities of VFMs. The findings confirm that VFMs serve as \textbf{Stronger} backbones, thereby establishing a significant benchmark in this field.

    \item We present a robust fine-tuning method, namely ``\textbf{Rein}", to parameter-efficiently harness VFMs. At its core, Rein consists of a set of learnable tokens, each directly linked to instances. With deliberate design, this linkage enables Rein to refine features at an instance-level within each layer. As a result, Rein reinforces the ability of VFMs in DGSS tasks, achieving this with \textbf{Fewer} trainable parameters while preserving the pre-trained knowledge.

    \item Comprehensive experiments across various DGSS settings demonstrate that Rein employs \textbf{Fewer} trainable parameters to effectively leverage \textbf{Stronger} VFMs for achieving \textbf{Superior} generalizability. This performance surpasses existing DGSS methods by a large margin. Notably, Rein is designed to integrate smoothly with existing plain vision transformers, improving their generalization ability and making training more efficient.
\end{itemize}
\setlength{\abovecaptionskip}{0.1cm}
\setlength{\belowcaptionskip}{0.1cm}
\begin{figure*}
    \centering
    \includegraphics[width=\linewidth]{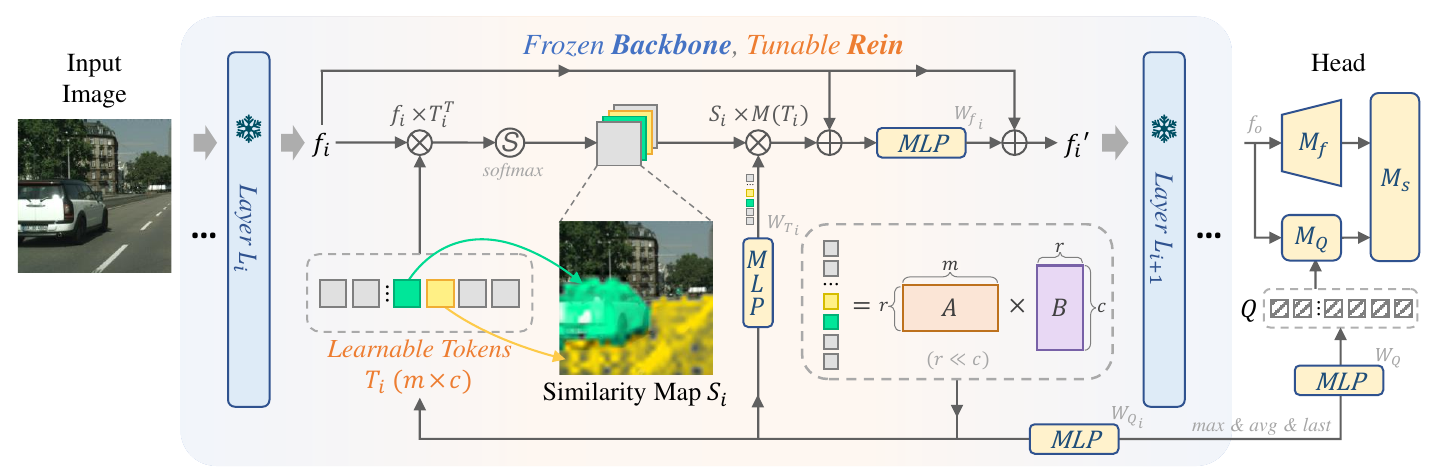}
    \caption{An overview of proposed Rein. Rein primarily consists of a collection of low-rank learnable tokens, denoted as $T=\{T_1,T_2,\ldots,T_N\}$. These tokens establish direct connections to distinct instances, facilitating instance-level feature refinement. This mechanism results in the generation of an enhancement feature map $f'_i=f_i+Rein(f_i)$ for each layer within backbone. All parameters of MLPs are layer-shared to reduce the number of parameters. $M_f$, $M_Q$, and $M_S$ are features module, queries module, and segmentation module, respectively. The notation $max~\&~avg~\&~last$ refers to the equation Eq.~(\ref{eq:link}) and Eq.~(\ref{eq:link2}).}
    \label{fig:framework}
    \vspace{-5mm}
\end{figure*}
\setlength{\abovecaptionskip}{0.1cm}
\setlength{\belowcaptionskip}{0.1cm}
\begin{table}[]
    \centering
    \resizebox{\linewidth}{!}{%
        \renewcommand\arraystretch{1.1}
        \setlength\tabcolsep{2.8pt}{
            \begin{tabular}{ll|c|cccc}
                \hline
                \rowcolor{white}
                \multirow{2}{*}{Backbone}                                                                 &
                \multirow{2}{*}{\begin{tabular}[c]{@{}l@{}}Fine-tune\\ Method\end{tabular}}               &
                \multirow{2}{*}{\begin{tabular}[c]{@{}c@{}}Trainable\\ Params$^*$\end{tabular}}           &
                \multicolumn{4}{c}{mIoU}                                                                                                                                                                               \\
                \cline{4-7}
                                                                                                          &        &          & Citys         & BDD           & Map           & Avg.                                   \\
                \hline
                \multirow{3}{*}{\begin{tabular}[c]{@{}l@{}}CLIP~\cite{CLIP}\\ (ViT-Large)\end{tabular}}   & Full   & 304.15M  & 51.3          & 47.6          & 54.3          & 51.1\cellcolor[HTML]{EFEFEF}          \\
                                                                                                          & Freeze & ~~~0.00M & 53.7          & 48.7          & 55.0          & 52.4\cellcolor[HTML]{EFEFEF}          \\
                                                                                                          & Rein   & ~~~2.99M & \textbf{57.1} & \textbf{54.7} & \textbf{60.5} & \textbf{57.4}\cellcolor[HTML]{EFEFEF} \\
                \hline
                \multirow{3}{*}{\begin{tabular}[c]{@{}l@{}}MAE~\cite{MAE}\\ (Large)\end{tabular}}         & Full   & 330.94M  & 53.7          & \textbf{50.8} & 58.1          & 54.2\cellcolor[HTML]{EFEFEF}          \\
                                                                                                          & Freeze & ~~~0.00M & 43.3          & 37.8          & 48.0          & 43.0\cellcolor[HTML]{EFEFEF}          \\
                                                                                                          & Rein   & ~~~2.99M & \textbf{55.0} & 49.3          & \textbf{58.6} & \textbf{54.3}\cellcolor[HTML]{EFEFEF} \\
                \hline
                \multirow{3}{*}{\begin{tabular}[c]{@{}l@{}}SAM~\cite{SAM}\\ (Huge)\end{tabular}}          & Full   & 632.18M  & 57.6          & 51.7          & 61.5          & 56.9\cellcolor[HTML]{EFEFEF}          \\
                                                                                                          & Freeze & ~~~0.00M & 57.0          & 47.1          & 58.4          & 54.2\cellcolor[HTML]{EFEFEF}          \\
                                                                                                          & Rein   & ~~~4.51M & \textbf{59.6} & \textbf{52.0} & \textbf{62.1} & \textbf{57.9}\cellcolor[HTML]{EFEFEF} \\
                \hline
                \multirow{3}{*}{\begin{tabular}[c]{@{}l@{}}EVA02~\cite{EVA,EVA02}\\ (Large)\end{tabular}} & Full   & 304.24M  & 62.1          & 56.2          & 64.6          & 60.9\cellcolor[HTML]{EFEFEF}          \\
                                                                                                          & Freeze & ~~~0.00M & 56.5          & 53.6          & 58.6          & 56.2\cellcolor[HTML]{EFEFEF}          \\
                                                                                                          & Rein   & ~~~2.99M & \textbf{65.3} & \textbf{60.5} & \textbf{64.9} & \textbf{63.6}\cellcolor[HTML]{EFEFEF} \\
                \hline
                \multirow{3}{*}{\begin{tabular}[c]{@{}l@{}}DINOV2~\cite{Dinov2}\\ (Large)\end{tabular}}   & Full   & 304.20M  & 63.7          & 57.4          & 64.2          & 61.7\cellcolor[HTML]{EFEFEF}          \\
                                                                                                          & Freeze & ~~~0.00M & 63.3          & 56.1          & 63.9          & 61.1\cellcolor[HTML]{EFEFEF}          \\
                                                                                                          & Rein   & ~~~2.99M & \textbf{66.4} & \textbf{60.4} & \textbf{66.1} & \textbf{64.3}\cellcolor[HTML]{EFEFEF} \\

                \hline
            \end{tabular}%
        }
    }
    \caption{Performance Comparison with the proposed \textbf{Rein across Multiple VFMs} as Backbones under the \textit{GTAV $\rightarrow$ Cityscapes (Citys) + BDD100K (BDD) + Mapillary (Map)} generalization setting. Mark $*$ denotes trainable parameters in backbones.}
    \label{tab:vfms_Rein}
    \vspace{-7mm}
\end{table}
\section{Related Works}
\label{related_work}
\noindent\textbf{Domain Generalized Semantic Segmentation.}
Domain Generalized Semantic Segmentation (DGSS) focuses on enhancing model generalizability. This field involves training models on a set of source domain to enhance their performance on distinct and unseen target domain. Various approaches~\cite{hgformer,SPC,MoDify,OCR,SAN-SAW,kamann2020increasing,ddb,dtp,parseall,trainone,coda} have been proposed to address this issue, with methods including splitting the learned features into domain-invariant and domain-specific components~\cite{dg:xu2022dirl,dg:tang2020selfnorm}, or employing meta-learning to train more robust models~\cite{PintheMem}. A standard scenario in DGSS is generalizing from one urban-scene dataset to another, for instance, from the synthetic GTAV~\cite{gtav} dataset to the real-world Cityscapes~\cite{cityscapes}. In this classic setting, certain techniques~\cite{choi2021robustnet,ibn,dg:switchable} have achieved notable performance through learning feature normalization/whitening schemes, while others~\cite{wildnet} have improved segmentation results through feature-level style transfer and the introduction of additional data. Additionally, strong data augmentation~\cite{advstyle,gtrltr,PASTA,fan2023towards} often simply and effectively enhances model robustness. However, most of previous DGSS methods generally utilize outdated backbones like ResNet~\cite{resnet}, VGGNet~\cite{vggnet}, MobileNetV2~\cite{mobilenet}, and ShuffleNetV2~\cite{shufflenet}, thereby leaving the efficacy of stronger Vision Foundation Models (VFMs) in DGSS relatively unexplored.

\noindent\textbf{Vision Foundation Models.}
The concept of a Foundation Model, initially introduced by Bommasani \etal~\cite{bommasani2022opportunities} in the field of Natural Language Processing (NLP), defined as ``the base models trained on large-scale data in a self-supervised or semi-supervised manner that can be adapted for several other downstream tasks". While models like the ViT~\cite{vit} and Swin Transformer~\cite{swin} have demonstrated excellent performance, the quest for a Vision Foundation Model (VFM) similar to their NLP counterparts is ongoing. This pursuit has yielded significant advancements with the advent of models such as CLIP~\cite{CLIP}, which learn high-quality visual representation by exploring contrastive learning with large-scale image text pairs; MAE~\cite{MAE}, utilizing a masked image modeling for learning latent image representations; SAM~\cite{SAM}, which develops a promptable model and pre-train it on a broad dataset for segmentation task; EVA02~\cite{EVA,EVA02}, which integrates Masked Image Modeling pre-training with CLIP's vision features; and DINOv2~\cite{Dinov2}, which is pretrained on extensive, curated datasets without explicit supervision. These VFMs have shown remarkable performance in downstream applications. Yet, a dedicated investigation into their performance in the specific context of DGSS tasks remains unexplored.

\noindent\textbf{Parameter-Efficient Fine-tuning.}
In the realm of NLP, parameter-efficient fine-tuning (PEFT) has achieved notable success by freezing most parameters of VFMs and fine-tuning a select few. Various approaches have been developed, such as BitFit~\cite{Bitfit}, which adjusts only the model's bias terms; Prompt-tuning~\cite{prompt_tuning}, introducing soft prompts to adapt frozen language models; Adapter-tuning~\cite{adapter}, adding lightweight modules to each transformer layer; and notably, LoRA~\cite{lora}, which injects trainable rank decomposition matrices into transformer layers, yielding significant influence.

The application of PEFT methods is also expanding into the field of computer vision~\cite{fahes2023simple,kumar2022fine}, with notable examples such as Visual Prompt Tuning (VPT)~\cite{vpt}, which prepends prompts into the input sequence of transformer layers; AdaptFormer~\cite{adaptformer}, replacing the MLP block in the transformer encoder with an AdaptMLP; LP-FT~\cite{kumar2022fine} find that fine-tuning can achieve worse accuracy than linear probing out-of-distribution; and Prompt-ICM~\cite{feng2023prompt}, applying large-scale pre-trained models to the task of image coding for machines. 
Contrasting with these methods, we aim to refine feature maps for each instance within an image, thereby achieving superior performance in the realm of DGSS.

\begin{table*}[tbp]
    \setlength{\abovecaptionskip}{0.1cm}
    \setlength{\belowcaptionskip}{0.1cm}
    \resizebox{\textwidth}{!}{%
\renewcommand\arraystretch{1.1}
\setlength\tabcolsep{2.5pt}{
\begin{tabular}{l|cccc|c|cccc|cccc|cccc|cccc|c}
\hline
\multirow{3}{*}{Target} &
  \multicolumn{5}{c|}{ACDC\cite{acdc} (test)} &
  \multicolumn{17}{c}{Cityscapes-C\cite{michaelis2019benchmarking} (level-5)} \\
  \cline{2-23}
 &
  \multirow{2}{*}{Night} &
  \multirow{2}{*}{Snow} &
  \multirow{2}{*}{Fog} &
  \multirow{2}{*}{Rain} &
  \multirow{2}{*}{All} &
  \multicolumn{4}{c|}{Blur} &
  \multicolumn{4}{c|}{Noise} &
  \multicolumn{4}{c|}{Digital} &
  \multicolumn{4}{c|}{Weather} &
  \multirow{2}{*}{Avg.} \\
  \cline{7-22}
 &
   &
   &
   &
   &
   &
  Motion &
  Defoc &
  Glass &
  Gauss &
  Gauss &
  Impul &
  Shot &
  Speck &
  Bright &
  Contr &
  Satur &
  JPEG &
  Snow &
  Spatt &
  Fog &
  Frost &
   \\
   \hline
HGFormer &
  52.7 &
  68.6 &
  69.9 &
  72.0 &
  67.2\cellcolor[HTML]{EFEFEF}&
   64.1 & 67.2 & 61.5 & 63.6 & \textbf{27.2} & \textbf{35.7} & \textbf{32.9} & 63.1 & 79.9 & 72.9 & 78.0 & 53.6 & 55.4 & 75.8 & 75.5 & 43.2&59.4\cellcolor[HTML]{EFEFEF}
  \\
Ours &
  \textbf{70.6} &
  \textbf{79.5} &
  \textbf{76.4} &
  \textbf{78.2} &
  \textbf{77.6}\cellcolor[HTML]{EFEFEF}&
   \textbf{68.5}
   &\textbf{71.7}
   &\textbf{69.7}
   &\textbf{68.7}
   &~6.2
   &23.0
   &13.1
   &\textbf{63.7}
   &\textbf{81.5}
   &\textbf{78.9}
   &\textbf{80.6}
   &\textbf{68.8}
   &\textbf{63.8}
   &\textbf{73.6}
   &\textbf{79.5}
   &\textbf{47.9}
   &\textbf{60.0}\cellcolor[HTML]{EFEFEF}
  \\
  \hline
\end{tabular}%
}
}
    \caption{Results on \textbf{Cityscapes $\rightarrow$ ACDC (test) and Cityscapes-C (level-5)}  datasets, utilizing a batch size of 8.}
    \label{tab:cityscapes-c+acdc}
    \vspace{-5mm}
\end{table*}
\setlength{\abovecaptionskip}{0.1cm}
\setlength{\belowcaptionskip}{0.1cm}
\begin{table}[tbp]
    \centering
    \resizebox{\linewidth}{!}{%
        \renewcommand\arraystretch{1.1}
        \setlength\tabcolsep{2.5pt}{
            \begin{tabular}{ll|c|cccc}
                \hline
                \rowcolor{white}
                \multirow{2}{*}{\begin{tabular}[c]{@{}l@{}}Backbone\end{tabular}}                           &
                \multirow{2}{*}{\begin{tabular}[c]{@{}l@{}}Fine-tune\\ Method\end{tabular}}                 &
                \multirow{2}{*}{\begin{tabular}[c]{@{}c@{}}Trainable\\ Params$^*$\end{tabular}}             &
                \multicolumn{4}{c}{mIoU}                                                                                                                                                                                                         \\
                \cline{4-7}
                                                                                                            &                                 &         & Citys         & BDD           & Map           & Avg.                                   \\
                \hline
                \multirow{10}{*}{\begin{tabular}[c]{@{}l@{}}EVA02\\(Large)\\~\cite{EVA,EVA02}\end{tabular}} & Full                            & 304.24M & 62.1          & 56.2          & 64.6          & 60.9\cellcolor[HTML]{EFEFEF}           \\
                                                                                                            & +AdvStyle~\cite{advstyle}       & 304.24M & 63.1          & 56.4          & 64.0          & 61.2\cellcolor[HTML]{EFEFEF}           \\
                                                                                                            & +PASTA~\cite{PASTA}             & 304.24M & 61.8          & 57.1          & 63.6          & 60.8\cellcolor[HTML]{EFEFEF}           \\
                                                                                                            & +GTR-LTR~\cite{gtrltr}           & 304.24M & 59.8          & 57.4          & 63.2          & 60.1\cellcolor[HTML]{EFEFEF}           \\
                \cline{2-7}
                                                                                                            & Freeze                          & ~~~0.00M & 56.5          & 53.6          & 58.6          & 56.2\cellcolor[HTML]{EFEFEF}           \\
                                                                                                            & +AdvStyle~\cite{advstyle}       & ~~~0.00M & 51.4          & 51.6          & 56.5          & 53.2\cellcolor[HTML]{EFEFEF}           \\
                                                                                                            & +PASTA~\cite{PASTA}             & ~~~0.00M & 57.8          & 52.3          & 58.5          & 56.2\cellcolor[HTML]{EFEFEF}           \\
                                                                                                            & +GTR-LTR~\cite{gtrltr}           & ~~~0.00M & 52.5          & 52.8          & 57.1          & 54.1\cellcolor[HTML]{EFEFEF}           \\
                                                                                                            & +LoRA~\cite{lora}               & ~~~1.18M & 55.5          & 52.7          & 58.3          & 55.5\cellcolor[HTML]{EFEFEF}           \\
                                                                                                            & +AdaptFormer~\cite{adaptformer} & ~~~3.17M & 63.7          & 59.9          & 64.2          & 62.6\cellcolor[HTML]{EFEFEF}           \\
                                                                                                            & +VPT~\cite{vpt}                 & ~~~3.69M & 62.2          & 57.7          & 62.5          & 60.8\cellcolor[HTML]{EFEFEF}           \\
                                                                                                            & +Rein (ours)                    & ~~~2.99M & \textbf{65.3} & \textbf{60.5} & \textbf{64.9} & \textbf{63.6} \cellcolor[HTML]{EFEFEF} \\
                \hline
                \multirow{10}{*}{\begin{tabular}[c]{@{}l@{}}DINOv2\\(Large)\\~\cite{Dinov2}\end{tabular}}   & Full                            & 304.20M & 63.7          & 57.4          & 64.2          & 61.7\cellcolor[HTML]{EFEFEF}           \\
                                                                                                            & +AdvStyle~\cite{advstyle}       & 304.20M & 60.8          & 58.0          & 62.5          & 60.4\cellcolor[HTML]{EFEFEF}           \\
                                                                                                            & +PASTA~\cite{PASTA}             & 304.20M & 62.5          & 57.2          & 64.7          & 61.5\cellcolor[HTML]{EFEFEF}           \\
                                                                                                            & +GTR-LTR~\cite{PASTA}           & 304.20M & 62.7          & 57.4          & 64.5          & 61.6\cellcolor[HTML]{EFEFEF}           \\
                \cline{2-7}
                                                                                                            & Freeze                          & ~~~0.00M & 63.3          & 56.1          & 63.9          & 61.1\cellcolor[HTML]{EFEFEF}           \\
                                                                                                            & +AdvStyle~\cite{advstyle}       & ~~~0.00M & 61.5          & 55.1          & 63.9          & 60.1\cellcolor[HTML]{EFEFEF}           \\
                                                                                                            & +PASTA~\cite{PASTA}             & ~~~0.00M & 62.1          & 57.2          & 64.5          & 61.3\cellcolor[HTML]{EFEFEF}           \\
                                                                                                            & +GTR-LTR~\cite{PASTA}           & ~~~0.00M & 60.2          & 57.7          & 62.2          & 60.0\cellcolor[HTML]{EFEFEF}           \\
                                                                                                            & +LoRA~\cite{lora}               & ~~~0.79M & 65.2          & 58.3          & 64.6          & 62.7\cellcolor[HTML]{EFEFEF}           \\
                                                                                                            & +AdaptFormer~\cite{adaptformer} & ~~~3.17M & 64.9          & 59.0          & 64.2          & 62.7\cellcolor[HTML]{EFEFEF}           \\
                                                                                                            & +VPT~\cite{vpt}                 & ~~~3.69M & 65.2          & 59.4          & 65.5          & 63.3\cellcolor[HTML]{EFEFEF}           \\
                                                                                                            & +Rein (ours)                    & ~~~2.99M & \textbf{66.4} & \textbf{60.4} & \textbf{66.1} & \textbf{64.3}\cellcolor[HTML]{EFEFEF}  \\
                \hline
            \end{tabular}%
        }
    }
    \caption{Performance Comparison of the proposed \textbf{Rein against other DGSS and PEFT methods} under the \textit{GTAV $\rightarrow$ Cityscapes (Citys) + BDD100K (BDD) + Mapillary (Map)} generalization setting. Mark $*$ denotes trainable parameters in backbones.}
    \label{tab:comparison}
    \vspace{-5mm}
\end{table}
\begin{table}[tbp]
    \setlength{\abovecaptionskip}{0.1cm}
    \setlength{\belowcaptionskip}{0.1cm}
    \centering
\renewcommand\arraystretch{1.1}
\setlength\tabcolsep{2.5pt}
\begin{tabular}{lc}
\hline
Source Domain                                                        & Cityscapes mIoU \\
\hline
GTAV                                                            & 66.4 \\
+Synthia                                                        & 68.1 \\
+UrbanSyn\cellcolor[HTML]{EFEFEF}                                                       & \textbf{78.4}\cellcolor[HTML]{EFEFEF} \\
\hline
\textbf{+1/16 of Cityscapes Training set}\cellcolor[HTML]{EFEFEF}                                                       & \textbf{82.5}\cellcolor[HTML]{EFEFEF} \\
\hline
\end{tabular}%
    \caption{\textbf{Synthetic data + 1/16 of Citys. $\rightarrow$ Citys.} val set.}
    \label{tab:dgss}
    \vspace{-5mm}
\end{table}
\section{Methods}
\label{sec:methods}
\subsection{Preliminary}
\label{sec:preliminary}
Driven by the motivation that \textbf{Leveraging Stronger pre-trained models and Fewer trainable parameters for Superior generalizability}, we choose to fine-tune VFMs with a reduced parameter set. A straightforward thought might involve a smaller decode head; however, this method merely acts as a passive receiver of feature maps from the backbone, lacking the flexibility to effectively adapt a frozen backbone for generating task-specific or scene-specific features. In contrast, we propose to embed a mechanism, named ``Rein", between the layers within the backbone. Rein actively refines and forwards the feature maps from each layer to the subsequent one. This approach allows us to more effectively utilize the powerful capabilities of VFMs, much like using rein to control a horse.

Given a pre-trained VFM with parameters $\Phi_{M}$, consisting of a sequence of layers $L_1,L_2,\ldots,L_N$, a decode head $\mathcal{H}$ parameterized by $\theta_{h}$, and the Rein strategy with parameters $\theta_{R}$, the optimization objective can be written as: 
\begin{equation}
\label{eq:peft_optimal}
\mathop{\arg\min}\limits_{\theta_{R},\theta_{h}} 
\sum_{i=1}^{N_d} \mathcal{L}oss(\mathcal{H}_{\theta_{h}}(\mathcal{F}_{\Phi_{M},\theta_{R}}(x_i)),y_i),
\end{equation}
where $x_i$ and $y_i$ denote the input image and its corresponding ground truth, respectively, and $N_d$ signifies the total number of samples. $\mathcal{F}_{\Phi_{M},\theta_{R}}$ represents the forward process of VFM after applying the Rein strategy.

\subsection{Core of Rein}
\label{sec:mainbody}
For simple implementation across different VFMs, we opt not to modify MLP weights at specific positions as described in the~\cite{adaptformer,lora}. Instead, our approach focuses on refining the output feature maps at each layer within the VFMs, as illustrated in Fig.~\ref{fig:framework}. Precisely, for the features $f_i$ produced by the $i$-th layer $L_i$, Rein produces enhanced feature maps for the next layer as follows:
\begin{equation}
    \begin{aligned}
        f_{1}&=L_{1}~(Embed(x))~~~~~~~~~~f_{1}\in \mathbb{R}^{n\times c}, \\ 
        f_{i+1}&=L_{i+1}(f_{i}+\Delta f_{i})~~~~~~~~~~i=1,2,\ldots,N-1, \\ 
        f_{out}&=f_N+\Delta f_N,  
    \end{aligned}    
    \label{eq:delta}
\end{equation}
where $f'_i=f_{i}+\Delta f_{i}$ symbolizes the refined feature map, $x$ is the input image, $Embed$ denotes the patch embedding layer in VFMs, $n$ represents the number of patches, $N$ denotes the number of layers, and $c$ is the dimensionality of $f_1,f_2,\ldots,f_N$. Note that the layers $L_1,L_2,\ldots,L_N$ are kept frozen, and our focus is on training an efficient module, Rein, to generate $\Delta f_{i}$ as follows:
\begin{equation}
    \Delta f_i=Rein(f_i)~~~~~~~~ \Delta f_i\in \mathbb{R}^{n\times c},i=1,2,\dots,N.
\end{equation}

In the context of DGSS, an ideal $\Delta f_{i}$ should assist VFMs to bridge two types of gaps. The first is gap in scene between pre-training dataset and target scene, exemplified by the contrast between ImageNet~\cite{imagenet} and urban-scene images~\cite{cityscapes,gtav}. The second is task divergence between pre-training and fine-tuning, such as the differences between masked image modeling and semantic segmentation tasks.

To establish this dual bridge, Rein starts with a set of learnable tokens $T=\{T_i\in \mathbb{R}^{m\times c}~|~i\in\mathbb{N},1\leq i\leq N\}$, where each token sequence $T_i$ is randomly initialized, and $m$ denotes the sequence length of $T_i$. Rein freezes the backbone and embeds knowledge learned from the fine-tuning dataset into these tokens, thereby bridging the gap in scene relative to the pre-training dataset. Moreover, considering the essential need in semantic segmentation to discern multiple instances within a single image, Rein implements an attention-inspired mechanism, which enables VFMs to make tailored adjustments to the features of distinct instances, thereby aiding VFMs in adapting to the differences between semantic segmentation and pre-training tasks. Specifically, Rein employs a dot-product operation to generate a similarity map $S_i$, which captures the associations between feature vectors in $f_i$ and the tokens in $T$:
\begin{equation}
\label{eq:dotproduct}
    S_i=f_i\times T_i^\text{T}~~~~~~~~ S_i\in \mathbb{R}^{n\times m},
\end{equation}
where $T_i$ represents the token sequence of the $i$-th layer, $m$ indicates the number of tokens in $T_i$. As $S$ quantitatively evaluates the relationships between various tokens and feature vectors, Rein can apply a softmax function to align each patch with a unique instance:
\begin{equation}
\label{eq:softmax}
S_i=Softmax(\frac{f_i\times T_i^\text{T}}{\sqrt{c}}).
\end{equation}

Leveraging the feature-to-token similarity map $S_i$, we can preliminarily estimates of $\Delta f_i$ using the equation:
\begin{equation}
\label{eq:obtain_delta}
\Delta \bar{f_i}=S_i(:,2:m)\times [~T_i(2:m)\times W_{T_i} + b_{T_i}],
\end{equation}
where $W_{T_i}$ and $b_{T_i}$ denote the weights and biases of a MLP, respectively. This MLP enables the transformation of $T_i$ across different feature spaces during the computation of $S_i$ and $\Delta \bar{f_i}$. Optionally, Rein can pre-calculate $T_i\times W_{T_i}+b_{T_i}$ to reduce inference time. 
The sum of $S_i$ equals one due to the softmax function; however, this can induce unneeded changes when all features are precise. To avoid this, $S_i(:,2:m)$ is designed to choose columns 2 to m of $S_i$, and $T_i(2:m)$ denotes the selection of rows 2 to m of $T_i$. This strategic selection allows models to sidestep unnecessary adjustments by assigning a high value to the first token and subsequently discarding it. This approach allows the sum of each row in $S_i$ to vary from $0$ to $1$, thus reducing the risk of inappropriate changes.

To enhance the flexibility in feature adjustment, Rein utilizes a MLP composed of $W_{f_i}$ and $b_{f_i}$ to produce the final feature modifications $\Delta f_i$:
\begin{equation}
\label{eq:finaldelta}
\Delta f_i=(\Delta \bar{f_i}+f_i) \times W_{f_i} +b_{f_i}.
\end{equation}

Benefiting from these instance-level $\Delta f_i$ adjustments, Rein is capable of generating diverse modifications for various categories within a single image. The details of Rein will be explained in the next section.

\subsection{Details of Rein}
\label{sec:further}
\noindent\textbf{Linking tokens to instances.} At the core of Rein, we establish an implicit yet effective linkage between tokens and instances, which has demonstrated notable performance, as detailed in Sec.~\ref{sec:experiments}. This connection is further reinforced by utilizing object queries, a key component in DETR\cite{detr}-style decode heads~\cite{maskformer,mask2former,segvit}, as intermediaries. These queries are empirically proven to establish a direct association with instances. Specifically, we generate layer-wise queries $Q_i$ from our learnable tokens $T_i$ via linear transformation:
\begin{equation}
\label{eq:link}
Q_i=T_i \times W_{Q_i}+b_{Q_i}~~~~~~~~Q_i\in\mathbb{R}^{m\times c'},
\end{equation}
where $W_{Q_i}$ and $b_{Q_i}$ signify the weights and biases, respectively, and $c'$ denotes the dimension of $Q_i$. However, due to the complexity arising from the large numbers of various layers in VFMs, transforming the diverse $Q_i$ into a single query $Q$ poses computational challenges. To address this, Rein computes both the maximal component $Q_{max} \in \mathbb{R}^{m \times c'}$ and the average component $Q_{avg} \in \mathbb{R}^{m \times c'}$ using the following equation:
\begin{equation}
    \begin{aligned}
        Q_{max}(j,k)&=\max_{i=1,2,\ldots,N}{Q_i(j,k)},\\ 
        Q_{avg}(j,k)&=\frac{1}{N}\sum_{i=1}^N{Q_i(j,k)}.
    \end{aligned}    
    \label{eq:max_avg}
\end{equation}
Subsequently, $Q$ is derived as:
\begin{align}
\label{eq:link2}
    Q&=Concat([Q_{max},Q_{avg},Q_N])\times W_Q + b_Q.
\end{align}
By mapping $T$ onto $Q$, which subsequently links to instances, Rein achieves enhanced performance with a marginal increase in parameters.

\noindent\textbf{Layer-shared MLP weights.} To address the redundancy of parameters in the layer-specific MLP weights, specifically $W_{T_i}$ in Eq.~(\ref{eq:obtain_delta}), $W_{f_i}$ in Eq.~(\ref{eq:finaldelta}), and $W_{Q_i}$ in Eq.~(\ref{eq:link}), which collectively contribute to a substantial trainable parameter count, we adopt a new strategy. Since the learnable $T_i$ is capable of producing distinct $\Delta f_i$ for each layer, we design the role of the MLP to primarily perform consistent linear transformations across different feature spaces for each layer within the backbone. To this end, we employ shared MLP weights across layers as outlined in the equations:
\begin{equation}
    \begin{aligned}
        &[W_{T_1},b_{T_1}]=[W_{T_2},b_{T_2}]=\ldots=[W_{T_N},b_{T_N}],\\ 
        &[W_{f_1},b_{f_1}]~=[W_{f_2},b_{f_2}]~~=\ldots=[W_{f_N},b_{f_N}],\\ 
        &[W_{Q_1},b_{Q_1}]=[W_{Q_2},b_{Q_2}]=\ldots=[W_{Q_N},b_{Q_N}].\label{eq:shareweight}
    \end{aligned}    
\end{equation}

\noindent\textbf{Low-rank token sequence.} Recognizing the potential for information overlap among diverse learnable tokens, such as the high similarity between tokens representing a car's headlight and a bicycle's light, Rein adopts a strategy to generate a low-rank token sequence $T$ as follows:
\begin{equation}
\label{eq:lowrank}
Ti=A_i\times B_i, ~~~~~~~A\in \mathbb{R}^{m\times r}, B\in \mathbb{R}^{r\times c},
\end{equation}
where $c$ denotes the dimension of $T_i$, $m$ is the length of sequence $T_i$, and $r$ represents the rank, with $r\ll c$. Here, matrices $A$ and $B$ are constructed as low-rank matrices. To reduce inference time, Rein can precompute and store $T$. By implementing this low-rank token sequence approach, Rein significantly reduces the number of parameter.
\section{Experiments}
\label{sec:experiments}
\subsection{Settings}
\label{sec:settings}
\noindent\textbf{Visual Foundation Models.}
To thoroughly assess the influence of Visual Foundation Models (VFMs) within the context of DGSS, we analyze five distinct VFMs, each with different training strategies and datasets. Our selection includes CLIP~\cite{CLIP}, a language-image pre-training model; MAE~\cite{MAE}, known for its masked pre-training approach; SAM~\cite{SAM}, which leverages a large-scale segmentation dataset; EVA02~\cite{EVA,EVA02} combines CLIP with masked image modeling; and DINOv2~\cite{Dinov2}, based on self-supervised pre-training with curated dataset. For balancing precision and efficiency, we mainly employ the ViT-Large architecture for these VFMs, except SAM, which utilizes a ViT-Huge image encoder, as described in its original paper~\cite{SAM}. We establish two fundamental baselines for VFMs: ``Full", where we fine-tune the entire network, and ``Freeze", in which all backbone parameters are fixed, with training solely on the segmentation head. More details about VFMs and PEFT methods are available in the supplementary material.

\begin{table*}[!htp]
    \setlength{\abovecaptionskip}{0.1cm}
    \setlength{\belowcaptionskip}{0.1cm}
    \resizebox{\linewidth}{!}{%
        \renewcommand\arraystretch{1.0}
        \setlength\tabcolsep{3.0pt}{
            \begin{tabular}{l|l|c|ccccccccccccccccccc|c}
            \hline
                Backbone                                                                                   & \begin{tabular}[l]{@{}l@{}}Fine-tune\\ Method\end{tabular} & \begin{tabular}[c]{@{}c@{}}Trainable\\ Params$^*$\end{tabular} & \multicolumn{1}{c}{road} & \multicolumn{1}{c}{side.} & \multicolumn{1}{c}{build.} & \multicolumn{1}{c}{wall} & \multicolumn{1}{c}{fence} & \multicolumn{1}{c}{pole} & \multicolumn{1}{c}{light} & \multicolumn{1}{c}{sign} & \multicolumn{1}{c}{vege} & \multicolumn{1}{c}{terr.} & \multicolumn{1}{c}{sky} & \multicolumn{1}{c}{pers.} & \multicolumn{1}{c}{rider} & \multicolumn{1}{c}{car} & \multicolumn{1}{c}{truck} & \multicolumn{1}{c}{bus} & \multicolumn{1}{c}{train} & \multicolumn{1}{c}{moto.} & \multicolumn{1}{c}{bicy.} & \multicolumn{1}{|l}{mIoU}    \\
                \hline
                \multirow{6}{*}{\begin{tabular}[c]{@{}l@{}}EVA02\\(Large)\\~\cite{EVA,EVA02}\end{tabular}} & Full                                                       & 304.24M\cellcolor[HTML]{EFEFEF}                                &89.3&46.9&89.9&47.7&45.6&\textbf{50.1}&56.8&42.2&88.8&\textbf{48.4}&89.9&\textbf{75.8}&\textbf{49.0}&90.5&45.3&69.2&55.9&44.4&55.1&62.2\\
                \cline{2-23}
                                                                                                           & Freeze                                                     & ~~~0.00M\cellcolor[HTML]{EFEFEF}                               &\textbf{93.1}\cellcolor[HTML]{FFEAB4}&52.7\cellcolor[HTML]{FFF3D7}&88.0\cellcolor[HTML]{FFFFFF}&47.4\cellcolor[HTML]{FFFFFF}&31.1\cellcolor[HTML]{FFFFFF}&41.7\cellcolor[HTML]{FFFFFF}&46.0\cellcolor[HTML]{FFFFFF}&39.6\cellcolor[HTML]{FFFFFF}&85.7\cellcolor[HTML]{FFFFFF}&41.4\cellcolor[HTML]{FFFFFF}&89.5\cellcolor[HTML]{FFFFFF}&67.5\cellcolor[HTML]{FFFFFF}&39.7\cellcolor[HTML]{FFFFFF}&89.0\cellcolor[HTML]{FFFFFF}&47.0\cellcolor[HTML]{FFFFFF}&72.8\cellcolor[HTML]{FFFFFF}&46.3\cellcolor[HTML]{FFFFFF}&19.2\cellcolor[HTML]{FFFFFF}&35.2\cellcolor[HTML]{FFFFFF}&56.5\cellcolor[HTML]{FFFFFF}\\
                                                                                                           & Rein-core                                                  & ~52.84M\cellcolor[HTML]{EFEFEF}                                &91.1\cellcolor[HTML]{FFFEFE}&53.8\cellcolor[HTML]{FFEDC0}&90.0\cellcolor[HTML]{FFEBBA}&50.3\cellcolor[HTML]{FFF8E9}&47.7\cellcolor[HTML]{FFEDC1}&46.6\cellcolor[HTML]{FFF5DE}&56.4\cellcolor[HTML]{FFECBE}&42.9\cellcolor[HTML]{FFF7E2}&87.8\cellcolor[HTML]{FFF6E2}&44.2\cellcolor[HTML]{FFF9EA}&90.4\cellcolor[HTML]{FFEDC2}&73.5\cellcolor[HTML]{FFEFC6}&44.2\cellcolor[HTML]{FFF7E2}&\textbf{91.8}\cellcolor[HTML]{FFEAB4}&\textbf{58.1}\cellcolor[HTML]{FFEAB4}&77.2\cellcolor[HTML]{FFFBF3}&57.3\cellcolor[HTML]{FFFCF6}&43.4\cellcolor[HTML]{FFEFC6}&57.3\cellcolor[HTML]{FFEEC4}&63.4\cellcolor[HTML]{FFF2D0}\\
                                                                                                           & $+$ Rein-link                                              & ~59.33M\cellcolor[HTML]{EFEFEF}                                &90.9\cellcolor[HTML]{FFFFFF}&48.5\cellcolor[HTML]{FFFFFF}&90.0\cellcolor[HTML]{FFEBBA}&52.6\cellcolor[HTML]{FFEBB9}&\textbf{49.4}\cellcolor[HTML]{FFEAB4}&49.1\cellcolor[HTML]{FFEAB4}&\textbf{57.2}\cellcolor[HTML]{FFEAB4}&39.8\cellcolor[HTML]{FFFEFE}&88.9\cellcolor[HTML]{FFECBC}&46.5\cellcolor[HTML]{FFEBB9}&\textbf{90.5}\cellcolor[HTML]{FFEAB4}&74.4\cellcolor[HTML]{FFEAB4}&44.0\cellcolor[HTML]{FFF7E4}&91.0\cellcolor[HTML]{FFF4D8}&52.3\cellcolor[HTML]{FFFAED}&80.7\cellcolor[HTML]{FFF4DA}&67.3\cellcolor[HTML]{FFF6DF}&44.3\cellcolor[HTML]{FFEEC2}&\textbf{60.3}\cellcolor[HTML]{FFEAB4}&64.1\cellcolor[HTML]{FFEFC7}\\
                                                                                                           & $+$ Rein-share                                             & ~~~5.02M\cellcolor[HTML]{EFEFEF}                               &92.7\cellcolor[HTML]{FFF0CC}&\textbf{54.3}\cellcolor[HTML]{FFEAB4}&90.0\cellcolor[HTML]{FFEBBA}&51.8\cellcolor[HTML]{FFF1CD}&48.6\cellcolor[HTML]{FFEBBA}&48.8\cellcolor[HTML]{FFEBB9}&55.3\cellcolor[HTML]{FFF0CB}&\textbf{45.0}\cellcolor[HTML]{FFEAB4}&88.9\cellcolor[HTML]{FFECBC}&46.7\cellcolor[HTML]{FFEAB4}&89.8\cellcolor[HTML]{FFFDF8}&73.7\cellcolor[HTML]{FFEEC2}&43.3\cellcolor[HTML]{FFF9EC}&90.6\cellcolor[HTML]{FFF8E6}&49.5\cellcolor[HTML]{FFFDFB}&81.1\cellcolor[HTML]{FFF3D6}&69.6\cellcolor[HTML]{FFF4D8}&41.7\cellcolor[HTML]{FFF1CE}&50.2\cellcolor[HTML]{FFF7E4}&63.4\cellcolor[HTML]{FFF2D0}\\
                                                                                                           & $+$ Rein-lora                                              & ~~~2.99M\cellcolor[HTML]{EFEFEF}                               &91.7\cellcolor[HTML]{FFFCF5}&51.8\cellcolor[HTML]{FFF8E6}&\textbf{90.1}\cellcolor[HTML]{FFEAB4}&\textbf{52.8}\cellcolor[HTML]{FFEAB4}&48.4\cellcolor[HTML]{FFECBB}&48.2\cellcolor[HTML]{FFEEC5}&56.0\cellcolor[HTML]{FFEEC3}&42.0\cellcolor[HTML]{FFFAF0}&\textbf{89.1}\cellcolor[HTML]{FFEAB4}&44.1\cellcolor[HTML]{FFF9EB}&90.2\cellcolor[HTML]{FFF4DA}&74.2\cellcolor[HTML]{FFEBB8}&47.0\cellcolor[HTML]{FFEAB4}&91.1\cellcolor[HTML]{FFF3D4}&54.5\cellcolor[HTML]{FFF5DC}&\textbf{84.1}\cellcolor[HTML]{FFEAB4}&\textbf{78.9}\cellcolor[HTML]{FFEAB4}&\textbf{47.2}\cellcolor[HTML]{FFEAB4}&59.4\cellcolor[HTML]{FFEBB9}&\textbf{65.3}\cellcolor[HTML]{FFEAB4}\\
                \hline
                \multirow{6}{*}{\begin{tabular}[c]{@{}l@{}}DINOv2\\(Large)\\~\cite{Dinov2}\end{tabular}}   & Full                                                       & 304.20M\cellcolor[HTML]{EFEFEF}                                &89.0&44.5&89.6&51.1&46.4&49.2&\textbf{60.0}&38.9&89.1&47.5&\textbf{91.7}&75.8&\textbf{48.2}&91.7&52.5&\textbf{82.9}&\textbf{81.0}&30.4&49.9&63.7\\
                \cline{2-23}
                                                                                                           & Freeze                                                     & ~~~0.00M\cellcolor[HTML]{EFEFEF}                               &92.1\cellcolor[HTML]{FFFBF3}&55.2\cellcolor[HTML]{FFFFFF}&90.2\cellcolor[HTML]{FFFFFF}&57.2\cellcolor[HTML]{FFF7E4}&48.5\cellcolor[HTML]{FFFFFF}&49.5\cellcolor[HTML]{FFFFFF}&56.7\cellcolor[HTML]{FFFFFF}&47.7\cellcolor[HTML]{FFF8E8}&89.3\cellcolor[HTML]{FFFFFF}&47.8\cellcolor[HTML]{FFFDF9}&91.1\cellcolor[HTML]{FFEAB4}&74.2\cellcolor[HTML]{FFFEFD}&46.7\cellcolor[HTML]{FFEFC5}&92.2\cellcolor[HTML]{FFF8E7}&62.6\cellcolor[HTML]{FFF5DD}&77.5\cellcolor[HTML]{FFFFFF}&47.7\cellcolor[HTML]{FFFFFF}&29.6\cellcolor[HTML]{FFFFFF}&47.2\cellcolor[HTML]{FFFFFF}&61.1\cellcolor[HTML]{FFFFFF}\\
                                                                                                           & Rein-core                                                  & ~52.84M\cellcolor[HTML]{EFEFEF}                                &92.4\cellcolor[HTML]{FFF9EA}&57.8\cellcolor[HTML]{FFFBF0}&90.6\cellcolor[HTML]{FFF1CF}&56.8\cellcolor[HTML]{FFFAF0}&50.7\cellcolor[HTML]{FFFBF1}&50.5\cellcolor[HTML]{FFFCF6}&57.5\cellcolor[HTML]{FFFDF9}&44.8\cellcolor[HTML]{FFFFFF}&\textbf{89.8}\cellcolor[HTML]{FFEAB4}&47.0\cellcolor[HTML]{FFFFFF}&91.1\cellcolor[HTML]{FFEAB4}&\textbf{75.9}\cellcolor[HTML]{FFEAB4}&47.2\cellcolor[HTML]{FFEAB4}&91.9\cellcolor[HTML]{FFFDFB}&60.1\cellcolor[HTML]{FFFDF9}&80.3\cellcolor[HTML]{FFF7E4}&59.8\cellcolor[HTML]{FFF9EA}&37.9\cellcolor[HTML]{FFF2D2}&52.3\cellcolor[HTML]{FFF3D4}&64.9\cellcolor[HTML]{FFF4D8}\\
                                                                                                           & $+$ Rein-link                                              & ~59.33M\cellcolor[HTML]{EFEFEF}                                &91.2\cellcolor[HTML]{FFFFFF}&55.5\cellcolor[HTML]{FFFEFE}&90.6\cellcolor[HTML]{FFF1CF}&55.6\cellcolor[HTML]{FFFFFF}&52.5\cellcolor[HTML]{FFF2D2}&51.1\cellcolor[HTML]{FFF8E8}&59.7\cellcolor[HTML]{FFEAB4}&45.1\cellcolor[HTML]{FFFEFE}&\textbf{89.8}\cellcolor[HTML]{FFEAB4}&47.1\cellcolor[HTML]{FFFEFE}&91.1\cellcolor[HTML]{FFEAB4}&75.8\cellcolor[HTML]{FFECBB}&47.1\cellcolor[HTML]{FFEBB7}&\textbf{92.6}\cellcolor[HTML]{FFEAB4}&\textbf{64.6}\cellcolor[HTML]{FFEAB4}&82.2\cellcolor[HTML]{FFEAB4}&65.5\cellcolor[HTML]{FFF2D2}&\textbf{40.4}\cellcolor[HTML]{FFEAB4}&52.7\cellcolor[HTML]{FFF1CD}&65.8\cellcolor[HTML]{FFEEC4}\\
                                                                                                           & $+$ Rein-share                                             & ~~~5.02M\cellcolor[HTML]{EFEFEF}                               &\textbf{93.5}\cellcolor[HTML]{FFEAB4}&\textbf{61.2}\cellcolor[HTML]{FFEAB4}&\textbf{90.7}\cellcolor[HTML]{FFEAB4}&57.7\cellcolor[HTML]{FFF2D1}&53.2\cellcolor[HTML]{FFEDC1}&\textbf{52.4}\cellcolor[HTML]{FFEAB4}&58.0\cellcolor[HTML]{FFFBF0}&\textbf{50.1}\cellcolor[HTML]{FFEAB4}&89.7\cellcolor[HTML]{FFF1CE}&\textbf{49.9}\cellcolor[HTML]{FFEAB4}&90.7\cellcolor[HTML]{FFFFFF}&74.8\cellcolor[HTML]{FFFAEF}&45.0\cellcolor[HTML]{FFFBF0}&91.7\cellcolor[HTML]{FFFFFF}&58.5\cellcolor[HTML]{FFFFFF}&80.1\cellcolor[HTML]{FFF8E8}&66.3\cellcolor[HTML]{FFF1CE}&36.9\cellcolor[HTML]{FFF5DC}&50.7\cellcolor[HTML]{FFF9EB}&65.8\cellcolor[HTML]{FFEEC4}\\
                                                                                                           & $+$ Rein-lora                                              & ~~~2.99M\cellcolor[HTML]{EFEFEF}                               &92.4\cellcolor[HTML]{FFF9EA}&59.1\cellcolor[HTML]{FFF6DF}&\textbf{90.7}\cellcolor[HTML]{FFEAB4}&\textbf{58.3}\cellcolor[HTML]{FFEAB4}&\textbf{53.7}\cellcolor[HTML]{FFEAB4}&51.8\cellcolor[HTML]{FFF1CF}&58.2\cellcolor[HTML]{FFF9EC}&46.4\cellcolor[HTML]{FFFDF8}&\textbf{89.8}\cellcolor[HTML]{FFEAB4}&49.4\cellcolor[HTML]{FFF0CB}&90.8\cellcolor[HTML]{FFFDFA}&73.9\cellcolor[HTML]{FFFFFF}&43.3\cellcolor[HTML]{FFFFFF}&92.3\cellcolor[HTML]{FFF5DD}&64.3\cellcolor[HTML]{FFECBB}&81.6\cellcolor[HTML]{FFEFC5}&70.9\cellcolor[HTML]{FFEAB4}&\textbf{40.4}\cellcolor[HTML]{FFEAB4}&\textbf{54.0}\cellcolor[HTML]{FFEAB4}&\textbf{66.4}\cellcolor[HTML]{FFEAB4}\\
                \hline
            \end{tabular}}}
    
    \caption{Ablation Study about Rein under \textit{Cityscapes $\rightarrow$ BDD100K} generalization in terms of mIoU. Components are sequentially incorporated. To better illustrate the gains contributed by each component, we employ varying shades of yellow to demonstrate the relative performance of the Freeze and Rein methods. The best results across all methods are \textbf{highlighted}.}
    \vspace{-14pt}
    \label{tab:class_iou}%
\end{table*}%

\begin{figure*}[htbp]
    \centering
    \setlength{\abovecaptionskip}{0.1cm}
    \setlength{\belowcaptionskip}{0.1cm}
    \includegraphics[width=\linewidth]{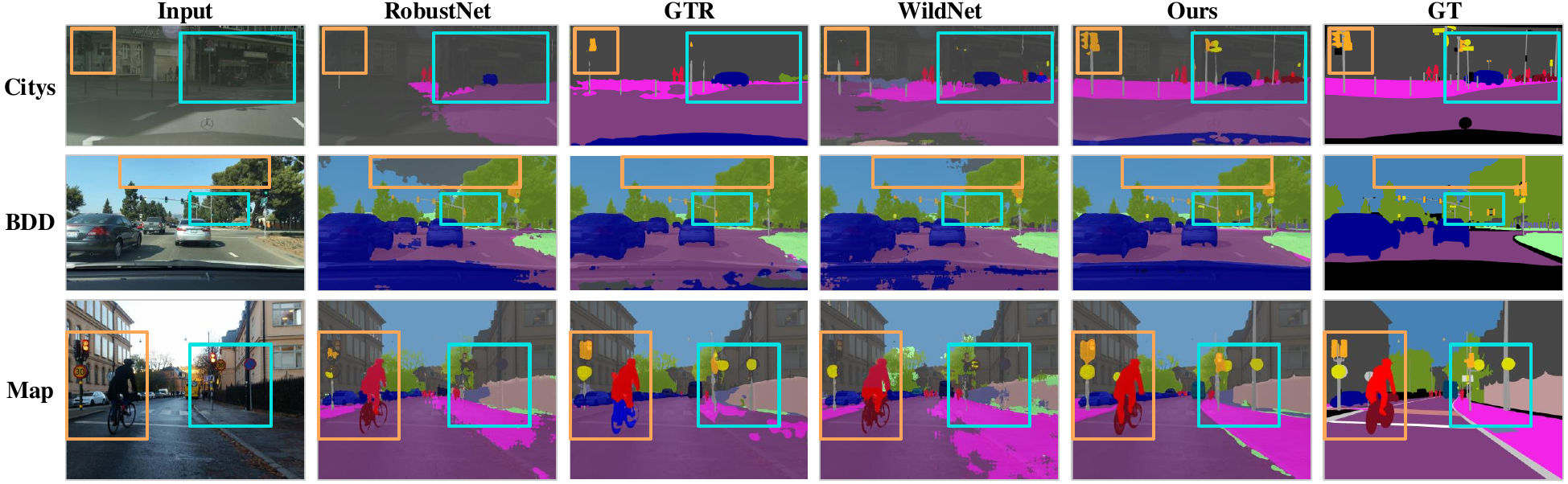}
    \caption{Qualitative Comparison under \textbf{\textit{GTAV $\rightarrow$ Cityscapes (Citys) + BDD100K (BDD) + Mapillary (Map)} generalization setting.}}
    \label{fig:qualitative}
    \vspace{-5mm}
\end{figure*}
\noindent\textbf{Datasets.} 
We evaluate VFMs and proposed methods on both real-world datasets (Cityscapes~\cite{cityscapes}, BDD100K~\cite{bdd100k}, Mapillary~\cite{mapillary}) and synthetic datasets (GTAV~\cite{gtav}, Synthia~\cite{SYNTHIA}, UrbanSyn~\cite{gómez2023one}). In detail, Cityscapes (denoted as Citys) is an autonomous driving dataset that contains 2975 training images and 500 validation images, each with the resolution of $2048\times1024$. BDD100K (shortened to BDD) and Mapillary (denoted by Map) offer 1,000 ($1280\times720$) and 2,000 ($1902\times1080$) validation images, respectively. GTAV, a synthetic dataset, presents 24,966 labeled images obtained from the game. Synthia, a synthetic dataset, provides 25,000 images created by photo-realistic rendering. UrbanSyn, a synthetic dataset consists of 7,539 images.

\noindent\textbf{Implementation details.}
We utilize the MMSegmentation ~\cite{mmseg2020} codebase for our implementation. For superior performance, mask2former~\cite{mask2former}, a widely-used segmentation head, is integrated with various VFMs that serve as the backbone. Additional experiments involving other decode heads are detailed in the supplementary material. For the training phase, the AdamW optimizer~\cite{adamw} is employed, setting the learning rate at 1e-5 for the backbone and 1e-4 for both the decode head and the proposed Rein. Aiming to efficient training process, we utilize a configuration of 40,000 iterations with a batch size of 4, and crop images to a resolution of $512\times512$. Our approach includes only basic data augmentation, following Mask2Former~\cite{mask2former}. Thanks to our streamlined training configuration and reduced number of trainable parameters, \textbf{Rein can fine-tune models like DINOv2-Large or EVA02-Large on a single RTX 3090Ti GPU within 12 hours} for superior generalization ability.

\subsection{Comparison with State-of-The-Art Methods}
In this section, we comprehensively evaluate Rein over five datasets within three generalization settings: \textit{GTAV $\rightarrow$ Citys + BDD + Map}, \textit{GTAV + Synthia $\rightarrow$ Citys + BDD + Map}, and \textit{Citys $\rightarrow$ BDD + Map}. Rein is benchmarked against state-of-the-art (SOTA) methods, which can be classified into two groups, including domain generalized semantic segmentation (DGSS) methods\cite{advstyle,PASTA,gtrltr,choi2021robustnet,PintheMem,SAN-SAW,wildnet,DIGA,SPC,ibn,drpc,hgformer}, and parameter-efficient fine-tuning (PEFT) approaches~\cite{vpt,lora,adaptformer}. 

\noindent\textbf{Investigation of various VFMs.}
Our analysis of VFMs and proposed Rein in the \textit{GTAV $\rightarrow$ Citys + BDD + Map} setting is presented in Tables~\ref{tab:inversitigate}~and~\ref{tab:vfms_Rein}. In this setup, models are fine-tuned using GTAV and evaluated on Cityscapes, BDD100K, and Mapillary. Note that, due to the fixed and relatively small number of trainable parameters in the decode head (20.6M), the count of trainable parameters presented in the tables are focused solely on the backbone and the PEFT module. Our results, as detailed in Table~\ref{tab:inversitigate}, indicate that frozen VFMs significantly outperform previous DGSS methods without specialized design. Moreover, as shown in Table~\ref{tab:vfms_Rein}, VFMs with full parameter fine-tuning exhibit enhanced performance relative to their frozen counterparts. Remarkably, Rein achieves even superior generalization capabilities, surpassing the full parameter fine-tuning  with merely an extra \textbf{1\%} of trainable parameters compared to the original backbone. Visual samples for qualitative comparison are given in Fig.~\ref{fig:qualitative}.

\noindent\textbf{Comparing Rein with SOTA.}
We conduct a comprehensive performance comparison of the proposed \textbf{Rein against existing DGSS and PEFT} methods under the \textit{GTAV $\rightarrow$ Citys + BDD + Map} setting, as detailed in Table~\ref{tab:comparison}. Owing to the robust feature extraction capabilities inherent in VFMs, DGSS methods, which typically enhance generalizability through strong data augmentation or consistency constraints, (\eg, AdvStyle, PASTA, and GTR), do not exhibit significant performance improvement. On the other hand, PEFT methods have demonstrated notable advancements. For instance, AdaptFormer outperforms the ``Freeze" baseline using EVA02 as the backbone, while VPT shows improved performance over ``Full" with DINOv2. Employing the same backbones (DINOv2 and EVA02), proposed Rein achieves superior performance and surpass previous DGSS and PEFT methods.

\noindent\textbf{Real-to-Real generalization of Rein.}
The generalization from one real-world dataset to others is pivotal for practical applications in the field. To this end, we conduct experiments under the \textbf{\textit{Citys $\rightarrow$ ACDC}}, \textit{Citys $\rightarrow$ Cityscapes-C}, and \textit{Citys $\rightarrow$ BDD + Map} generalization setting. As shown in Table~\ref{tab:cityscapes-c+acdc} and \ref{tab:c2bm}, Rein, when coupled with the DINOv2-Large, demonstrates superior performance across all datasets. This underscores the effectiveness of Rein in generalizing to diverse real-world scenarios.

\noindent\textbf{Synthetic-to-real generalization of Rein.}
As Tab.~\ref{tab:dgss} illustrates, trained on \textbf{synthetic} \textit{UrbanSyn+GTAV+Synthia} datasets, Rein achieved a \textbf{78.4\% mIoU} on the Cityscapes validation set. Further improvement is possible with additional synthetic data and higher-quality images generated by diffusion models, like \cite{benigmim2023collaborating}. This result can also be a valuable pre-trained weight for data-efficient training, \textbf{reaching an $82.5\%$ mIoU with 1/16 of Cityscapes training set}. This is a significant performance for semi-supervised semantic segmentation.

\noindent\textbf{More backbones.} We extend our analysis to integrating Rein with Convolutional Networks, such as ResNet and ConvNeXt, and smaller scale architectures like DINOv2-S/B. As shown in Table~\ref{tab:backbones}, our findings reveal that Rein exhibits remarkable performance with diverse backbones.

\subsection{Ablation Studies and Analysis}
We conduct extensive ablation studies within two settings: \textit{GTAV $\rightarrow$ Citys} and \textit{GTAV $\rightarrow$ Citys + BDD + Map}.

\noindent\textbf{Analysis of the key components.} 
Table~\ref{tab:class_iou} is dedicated to thoroughly examining the effectiveness of each component within Rein. In the \textit{GTAV $\rightarrow$ Citys} generalization setting, we sequentially incorporate different components of Rein and assess their impact. Interestingly, we observe that the ``Freeze" occasionally exhibit better recognition for specific categories, \eg, `road, sidewalk', compared to the ``Full". This suggests that VFMs lose some pre-training knowledge during fine-tuning, and ``Freeze" helps to prevent. Similarly, our methods mitigate this knowledge forgetting. Furthermore, our methods show improved recognition capabilities for the majority of the 19 categories. For example, in recognizing `wall, motorcycle, bicycle', our approach significantly outperforms both the ``Full" and ``Freeze" baselines.

Overall, ``Rein-core" boosts the average performance across 19 classes. Furthermore, ``Rein-link" further boosts accuracy for certain objects, including `car, bus, train, motorcycle', especially for DINOv2. Employing layer-shared MLP weights and low-rank token sequence efficiently reduces the number of trainable parameters and positively influences the performance of the model.

\setlength{\abovecaptionskip}{0.1cm}
\setlength{\belowcaptionskip}{0.1cm}
\begin{table}[tbp]
    \centering
    \resizebox{\linewidth}{!}{%
        \renewcommand\arraystretch{1.0}
        \setlength\tabcolsep{3.5pt}{
            \begin{tabular}{l|c|c|ccc}
                \hline
                \multirow{2}{*}{Methods} & \multirow{2}{*}{Backbone}  & \multirow{2}{*}{\begin{tabular}[c]{@{}c@{}}Trainable\\Parameters$^*$\end{tabular}} & \multicolumn{3}{c}{mIoU}                                                                   \\
                \cline{4-6}
                                         &                            &                                                                                    & \multicolumn{1}{c}{BDD}  & \multicolumn{1}{c}{Map} & \multicolumn{1}{c}{Avg.}              \\
                \hline
                IBN~\cite{ibn}           & ResNet50~\cite{resnet}     & ~~23.58M                                                                            & 48.6                     & 57.0                    & 52.8\cellcolor[HTML]{EFEFEF}          \\
                DRPC~\cite{drpc}         & ResNet50~\cite{resnet}     & ~~23.58M                                                                            & 49.9                     & 56.3                    & 53.1\cellcolor[HTML]{EFEFEF}          \\
                GTR~\cite{gtrltr}        & ResNet50~\cite{resnet}     & ~~23.58M                                                                            & 50.8                     & 57.2                    & 54.0\cellcolor[HTML]{EFEFEF}          \\
                SAN-SAW~\cite{SAN-SAW}   & ResNet50~\cite{resnet}     & ~~23.58M                                                                            & 53.0                     & 59.8                    & 56.4\cellcolor[HTML]{EFEFEF}          \\
                WildNet~\cite{wildnet}   & ResNet101~\cite{resnet}    & ~~42.62M                                                                            & 50.9                     & 58.8                    & 54.9\cellcolor[HTML]{EFEFEF}          \\
                HGFormer~\cite{hgformer} & Swin-L~\cite{swin}                 & 196.03M                                                                            & 61.5                     & 72.1                    & 66.8\cellcolor[HTML]{EFEFEF}          \\
                \hline
                Freeze                   & EVA02-L~\cite{EVA02}   & ~~~0.00M                                                                                  & 57.8                     & 63.8                    & 60.8\cellcolor[HTML]{EFEFEF}          \\
                Rein (Ours)              & EVA02-L~\cite{EVA02}   & ~~~2.99M                                                                            & 64.1                     & 69.5                    & 66.8\cellcolor[HTML]{EFEFEF}          \\
                \hline
                Freeze                   & DINOv2-L~\cite{Dinov2} & ~~~0.00M                                                                                  & 63.4                     & 69.7                    & 66.7\cellcolor[HTML]{EFEFEF}          \\
                Rein (Ours)              & DINOv2-L~\cite{Dinov2} & ~~~2.99M                                                                            & \textbf{65.0}            & \textbf{72.3}           & \textbf{68.7}\cellcolor[HTML]{EFEFEF} \\
                \hline
            \end{tabular}
        }
    }
    \caption{Performance Comparison of the \textbf{Rein against other DGSS methods} under \textit{Cityscapes $\rightarrow$ BDD100K (BDD) +Mapillary (Map)} generalization. The best results are \textbf{highlighted}.}
    \label{tab:c2bm}
    \vspace{-4mm}
\end{table}
\begin{table}[tbp]
    \setlength{\abovecaptionskip}{0.1cm}
    \setlength{\belowcaptionskip}{0.1cm}
    \centering
    \renewcommand\arraystretch{1.1}
\setlength\tabcolsep{2pt}
\begin{tabular}{l|ccc|cc}
\hline
Avg. mIoU                   & \begin{tabular}[c]{@{}c@{}}ResNet\\(50)\end{tabular} & \begin{tabular}[c]{@{}c@{}}ResNet\\(101)\end{tabular} & \begin{tabular}[c]{@{}c@{}}ConvNeXt\\(Large)\end{tabular} & \begin{tabular}[c]{@{}c@{}}DINOv2\\(S)\end{tabular} & \begin{tabular}[c]{@{}c@{}}DINOv2\\(B)\end{tabular} \\ \hline
Full                       & 38.9     & 46.1      & 52.2       & 51.8  &  56.7 \\ 
Ours\cellcolor[HTML]{EFEFEF}                      & \textbf{46.6}\cellcolor[HTML]{EFEFEF}     & \textbf{46.3}\cellcolor[HTML]{EFEFEF}      & \textbf{55.5}\cellcolor[HTML]{EFEFEF}       & \textbf{55.7}\cellcolor[HTML]{EFEFEF}  & \textbf{59.1}\cellcolor[HTML]{EFEFEF}
\\ \hline
\end{tabular}%
    \caption{Results for \textbf{ConvNets and smaller backbones}.}
    \label{tab:backbones}
    \vspace{-5mm}
\end{table}

\begin{figure}[tbp]
    \centering
    \setlength{\abovecaptionskip}{0.1cm}
    \setlength{\belowcaptionskip}{0.1cm}
    \includegraphics[width=\linewidth]{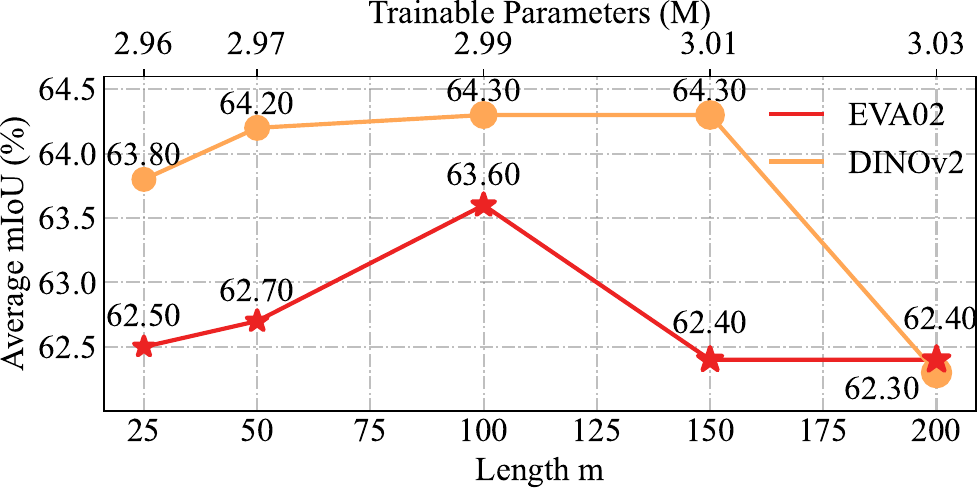}
    \caption{Ablation study on token length $m$.}
    \label{fig:ablation_length}
    \vspace{-4mm}
\end{figure}

\noindent\textbf{Study on token length $m$.}
The core component of Rein is learnable tokens $T\in\mathbb{R}^{m\times c}$. We explored various lengths $m$ for the token sequence, ranging from 25 to 200. As demonstrated in Fig.~\ref{fig:ablation_length}, models with $m=100$ and $m=150$ both achieve a strong mIoU of 64.3\%. We ultimately selected $m=100$ as the most suitable parameter.

\noindent\textbf{Study on rank $r$.}
As shown in Table~\ref{tab:ablation_lora}, we turn attention to the effect of rank $r$ on model performance. With DINOv2 as the backbone, the optimal results are observed at $r=16$ and $r=32$. Consequently, unlike LoRA~\cite{lora}, we opt for a comparatively higher value of $r=16$ for our model.

\noindent\textbf{Speed, memory, and storage.} 
For practical applications, training speed, GPU memory usage, and model storage requirements are crucial. As shown in Table~\ref{tab:speed_mem_storage}, compared to ``Full" baseline, proposed Rein improves training speed and reduces GPU memory usage. A significant advantage of Rein is that models trained under different settings can share the same backbone parameters. This means that for switch in diverse tasks and settings, we can only store and swap the rein weights (0.01GB) and head weights (0.08GB), rather than all parameters.
\setlength{\abovecaptionskip}{0.1cm}
\setlength{\belowcaptionskip}{0.1cm}
\begin{table}[tbp]
    \centering
        \renewcommand\arraystretch{1.0}
        \setlength\tabcolsep{3.5pt}{
            \begin{tabular}{ll|ccccc}
                \hline
                \multicolumn{2}{l|}{Rank $r$}                                                          & 4                            & 8                            & 16                           & 32                           & 64                                                          \\
                \multicolumn{2}{l|}{Params}                                                              & 2.67M                        & 2.77M                        & 2.99M                        & 3.42M                        & 4.28M                                                       \\
                \hline
                \multirow{4}{*}{\begin{tabular}[c]{@{}l@{}}DINOv2\\(Large)\\~\cite{Dinov2}\end{tabular}} & Citys                        & 65.8                         & 66.1                         & 66.4                         & 66.1                         & 66.4                         \\
                                                                                                         & BDD                          & 60.2                         & 60.3                         & 60.4                         & 60.7                         & 61.0                         \\
                                                                                                         & Map                          & 65.2                         & 65.1                         & 66.1                         & 65.9                         & 65.0                         \\
                \cline{2-7}
                                                                                                         & Avg.\cellcolor[HTML]{EFEFEF} & 63.7\cellcolor[HTML]{EFEFEF} & 63.9\cellcolor[HTML]{EFEFEF} & \textbf{64.3}\cellcolor[HTML]{EFEFEF} & \textbf{64.3}\cellcolor[HTML]{EFEFEF} & 64.1\cellcolor[HTML]{EFEFEF} \\

                \hline
            \end{tabular}%
        }
    \caption{Ablation study on lora dim $r$.}
    \label{tab:ablation_lora}
    \vspace{-4mm}
\end{table}
\begin{table}[!tbp]
    \centering
    \setlength{\abovecaptionskip}{0.1cm}
    \setlength{\belowcaptionskip}{0.1cm}
    \renewcommand\arraystretch{1.0}
        \setlength\tabcolsep{3.5pt}{
        \begin{tabular}{l|l|ccc}
            \hline
            \multirow{2}{*}{VFMs}                                                         &
            \multirow{2}{*}{Method}                                                       &
            \multirow{2}{*}{\begin{tabular}[c]{@{}c@{}}Training\\Time\end{tabular}} &
            \multirow{2}{*}{\begin{tabular}[c]{@{}c@{}}GPU\\Memory\end{tabular}}     &
            \multirow{2}{*}{\begin{tabular}[c]{@{}c@{}}Storage\end{tabular}}                                    \\
                                                                                          &      &      &      &      \\
            \hline
            \multirow{2}{*}{\begin{tabular}[c]{@{}c@{}} DINOv2\\(Large)\end{tabular}}     & Full & 11.2 h & 14.7 GB & 1.22 GB \\
                                                                                          & Rein & ~~9.5 h & 10.0 GB & 1.23 GB \\
            \hline
        \end{tabular}%
    }
    
    \caption{Training Time, GPU Memory, and Storage.}
    \label{tab:speed_mem_storage}
    \vspace{-5mm}
\end{table}
\vspace{-0.5mm}
\section{Conclusions}
\vspace{-0.5mm}
\label{conclusions}
In this paper, we assess and harness Vision Foundation Models (VFMs) in the context of DGSS. Driven by the motivation that \textbf{Leveraging Stronger pre-trained models and Fewer trainable parameters for Superior generalizability}, we first investigate the performance of VFMs under diverse DGSS settings. Subsequently, we introduce a robust fine-tuning approach, namely \textbf{Rein}, to parameter-efficiently harness VFMs for DGSS. With a fewer trainable parameters, Rein significantly enhance generalizability of VFMs, outperforming SOTA methods by a large margin. Rein can be seamlessly integrated as a plug-and-play adapter for existing VFMs, improving generalization with efficient training. Extensive experiments demonstrate the substantial potential of VFMs in the DGSS field, validating the effectiveness of proposed Rein in harnessing VFMs for DGSS.
\section{Acknowledgements}
This work was supported in part by the Anhui Provincial Key Research and Development Plan 202304a05020072, in part by the Postdoctoral Fellowship Program of CPSFGZB20230713, and in part by the National Natural Science Foundation of China under Grant 61727809.

\twocolumn[{%
    \renewcommand\twocolumn[1][]{#1}%
    \maketitlesupplementary
    \setlength{\abovecaptionskip}{0.1cm}
    \setlength{\belowcaptionskip}{0.1cm}
	\begin{center}
		\centering
        \vspace{-0.6cm}
        \begin{thry_algorithm}
        \end{thry_algorithm}
    \includegraphics[width=\linewidth]{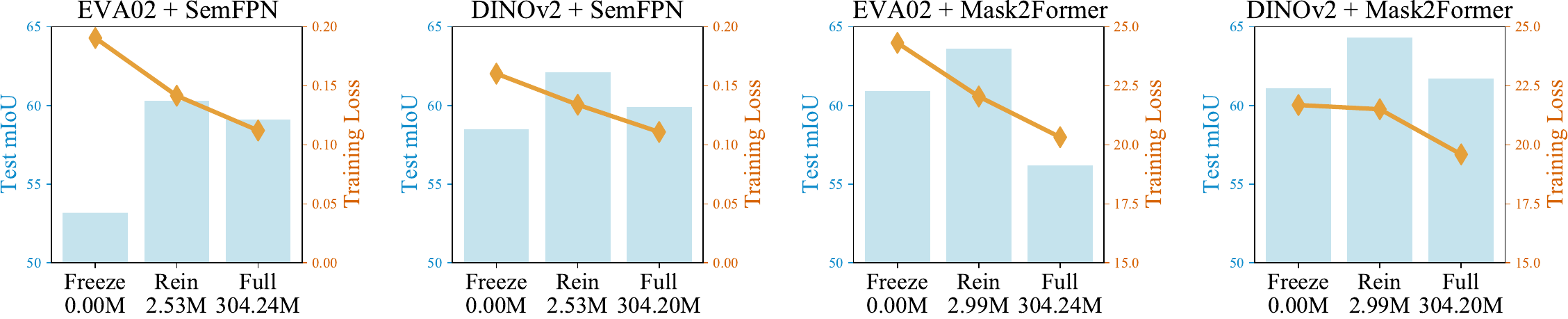}
    \captionof{figure}{The curves of training loss and test metrics display consistent trends across different VFMs and decode heads: intuitively, as trainable parameters increase from $0.00M (Freeze)\rightarrow 2.53M (Rein) \rightarrow 304.24M (Full)$, the training loss monotonically decreases, indicating that a greater number of trainable parameters indeed better fit the training dataset. However, the test metrics on the target dataset initially rise and then fall, forming an inverted U-shape. This pattern suggests that the ``Full" baseline overfits the training data, leading to diminished test performance. These findings are aligned with our motivation that \textbf{Leveraging Stronger pre-trained models and Fewer trainable parameters for Superior generalizability}. The blue bar charts in the figure represent the average mIoU tested on the Cityscapes, BDD100K, and Mapillary datasets, while the yellow line denotes the training loss during fine-tuning on GTAV dataset.}
    \label{fig:mxx}
    \end{center}     
}]
\section{Fewer Trainable Parameters}
Classical neural network theory~\cite{delemma,hastie2009elements} points out that as model capacity increases, the empirical risk (or training risk) monotonically decreases, indicating an improved fit to training data. Conversely, the true risk (or test risk) typically exhibits a ``U-shaped" curve, initially decreasing and then increasing, a phenomenon known as overfitting. From a modern viewpoint, the scaling law~\cite{scaling} suggests that on a smaller fixed dataset, performance stops to improve as model parameters increase, leading to overfitting.

In the majority of general tasks, the practice of early-stopping, based on evaluation data, can partly mitigate overfitting. However, in the field of domain generalization, the unknown test data distribution makes acquiring a valid evaluation dataset unavailable. Moreover, fine-tuning datasets are often smaller compared to ImageNet~\cite{imagenet} or LVD-142M~\cite{Dinov2}. Hence, employing fewer trainable parameters emerges as a strategic approach to mitigate overfitting.

In our main paper, extensive experiments comprehensively demonstrate Rein's pivotal role in enhancing the generalization capabilities of VFMs. This enhancement may be attributed to two factors: 1) Rein's improved fitting capability for VFMs, ensuring better alignment with training data; 2) Rein's reduction of overfitting in VFMs during fine-tuning on smaller datasets, thus exhibiting enhanced generalization in testing. To delve into this, we analyze and compare the average training loss in the final 1000 iterations of the fine-tuning phase and their corresponding test metrics for various VFMs and decode heads.

Fig.~\ref{fig:mxx} showcases a consistent trend across four different configurations. As trainable parameters increase from $0.00M (Freeze)\rightarrow 2.53M (Rein) \rightarrow 304.24M (Full)$, the training loss monotonically decreases. However, the test metrics on the target dataset peak with Rein, which employs 2.53 million parameters and incurs a sub-optimal training loss. In contrast, the ``Full" baseline, despite recording the lowest training loss, only achieves sub-optimal test performance, a clear indicator of overfitting when compared to other setups. This observation aligns with the conclusions in ~\cite{delemma,scaling}, supporting ours observation that \textbf{leveraging Stronger pre-trained models and Fewer trainable parameters can lead to Superior generalizability}. 
\begin{table}[htbp]
    \centering
\renewcommand\arraystretch{1.1}
\setlength\tabcolsep{2.5pt}
\begin{tabular}{lc}
\hline
Source Domain                                                        & Cityscapes mIoU \\
\hline
GTAV                                                            & 66.4 \\
+Synthia                                                        & 68.1 \\
+UrbanSyn\cellcolor[HTML]{EFEFEF}                                                       & \textbf{78.4}\cellcolor[HTML]{EFEFEF} \\
\hline
\textbf{+1/16 of Cityscapes Training set}\cellcolor[HTML]{EFEFEF}                                                       & \textbf{82.5}\cellcolor[HTML]{EFEFEF} \\
\hline
\end{tabular}%
    \caption{Results on Cityscapes validation set.}
    \label{tab:dgss}
\end{table}
\section{Value of synthetic data}
As Tab.~\ref{tab:dgss} illustrates, trained on synthetic \textit{UrbanSyn~\cite{gómez2023one}+GTA+Synthia} datasets, Rein achieved a \textbf{78.4\% mIoU} on the Cityscapes validation set. Further improvement is possible with additional synthetic data and higher-quality images generated by diffusion models, like \cite{benigmim2023collaborating}. This result can also be a valuable pre-trained weight for data-efficient training, reaching an $82.5\%$ mIoU with 1/16 of Cityscapes training set. This is a significant performance for semi-supervised semantic segmentation.

\section{Ablation on decode head}
Our experiments on Rein employ the Mask2Former~\cite{mask2former} decode head, which shares structures or core concepts with numerous methods in dense prediction tasks~\cite{LiDARmask2former,segvit,maskdino,detr,maskformer}. The universality of Mask2Former highlights the significance of our findings for a range of segmentation tasks, including instance and panoptic segmentation. Furthermore, to demonstrate Rein's effectiveness in enhancing backbone generalization and its robustness across various decode heads, we conduct supplementary experiments using the popular SemFPN decode head~\cite{semfpn}, in the \textit{GTAV$\rightarrow$ Cityscapes + BDD100K + Mapillary} setting.

As shown in Table~\ref{tab:vfms_Rein_semfpn}, Rein surpasses the ``Full" and ``Freeze" baselines, employing 2.53 million trainable parameters within the backbone, while the SemFPN decode head comprises 1.63 million parameters. Owing to the absence of object queries in SemFPN, the ``linking tokens to instance" mechanism, described in Sec.3.3, is not utilized, resulting in a reduction of Rein's trainable parameters from 2.99 million to 2.53 million. When compared to the complete Rein configuration using the Mask2Former, using SemFPN achieves sub-optimal performance, evident in the 64.3\% mIoU reported in Table~2 and 62.1\% mIoU in Table~9, both implemented with DINOv2-Large. As shown in Table~\ref{tab:benefits}, the Mask2Former brings the 11.7\% mIoU for ResNet101.These findings guide our decision to focus on experiments involving Mask2Former in the main paper.
\section{Ablation on EVA02}
\setlength{\abovecaptionskip}{0.1cm}
\setlength{\belowcaptionskip}{0.1cm}
\begin{table}[tbp]
    \centering
        \renewcommand\arraystretch{1.0}
        \setlength\tabcolsep{3.5pt}{
            \begin{tabular}{ll|ccccc}
                \hline
                \multicolumn{2}{l|}{Rank $r$}                                                          & 4                            & 8                            & 16                           & 32                           & 64                                                          \\
                \multicolumn{2}{l|}{Params}                                                              & 2.67M                        & 2.77M                        & 2.99M                        & 3.42M                        & 4.28M                                                       \\
                \hline
                \multirow{4}{*}{\begin{tabular}[c]{@{}l@{}}EVA02\\(Large)\\~\cite{EVA02}\end{tabular}}   & Citys                        & 62.6                         & 63.5                         & 65.3                         & 63.8                         & 63.4                         \\
                                                                                                         & BDD                          & 58.5                         & 58.9                         & 60.5                         & 60.5                         & 60.2                         \\
                                                                                                         & Map                          & 63.7                         & 63.8                         & 64.9                         & 64.5                         & 64.3                         \\
                \cline{2-7}
                                                                                                         & Avg.\cellcolor[HTML]{EFEFEF} & 61.6\cellcolor[HTML]{EFEFEF} & 62.1\cellcolor[HTML]{EFEFEF} & \textbf{63.6}\cellcolor[HTML]{EFEFEF} & 62.9\cellcolor[HTML]{EFEFEF} & 62.7\cellcolor[HTML]{EFEFEF} \\
                \hline

            \end{tabular}%
        }
    \caption{Ablation study on lora dim $r$.}
    \label{tab:ablation_lora}
    \vspace{-4mm}
\end{table}
\begin{table}[!tbp]
    \centering
    \setlength{\abovecaptionskip}{0.1cm}
    \setlength{\belowcaptionskip}{0.1cm}
    \renewcommand\arraystretch{1.0}
        \setlength\tabcolsep{3.5pt}{
        \begin{tabular}{l|l|ccc}
            \hline
            \multirow{2}{*}{VFMs}                                                         &
            \multirow{2}{*}{Method}                                                       &
            \multirow{2}{*}{\begin{tabular}[c]{@{}c@{}}Training\\Time\end{tabular}} &
            \multirow{2}{*}{\begin{tabular}[c]{@{}c@{}}GPU\\Memory\end{tabular}}     &
            \multirow{2}{*}{\begin{tabular}[c]{@{}c@{}}Storage\end{tabular}}                                    \\
                                                                                          &      &      &      &      \\
            \hline
            \multirow{2}{*}{\begin{tabular}[c]{@{}c@{}} EVA02\\(Large)\end{tabular}}      & Full & 11.8 h & 15.9 GB & 1.22 GB \\
                                                                                          & Rein & 10.5 h & 12.5 GB & 1.23 GB \\
            \hline
            \hline
        \end{tabular}%
    }
    
    \caption{Training Time, GPU Memory, and Storage.}
    \label{tab:speed_mem_storage}
    \vspace{-4mm}
\end{table}
\noindent\textbf{Study on rank $r$}
As shown in Table~\ref{tab:ablation_lora}, with EVA02 as the backbone, the optimal results are observed at $r=16$.

\noindent\textbf{Speed, memory, and storage.} 
As shown in Table~\ref{tab:speed_mem_storage}, compared to ``Full" baseline, proposed Rein improves training speed and reduces GPU memory usage.

\begin{table}[tbp]
    \centering
    \resizebox{\linewidth}{!}{%
        \renewcommand\arraystretch{1.1}
        \setlength\tabcolsep{2.8pt}{
            \begin{tabular}{ll|c|cccc}
                \hline
                \rowcolor{white}
                \multirow{2}{*}{Backbone}                                                                 &
                \multirow{2}{*}{\begin{tabular}[c]{@{}l@{}}Fine-tune\\ Method\end{tabular}}               &
                \multirow{2}{*}{\begin{tabular}[c]{@{}c@{}}Trainable\\ Params$^*$\end{tabular}}           &
                \multicolumn{4}{c}{mIoU}                                                                                                                                                                              \\
                \cline{4-7}
                                                                                                          &        &          & Citys         & BDD           & Map           & Avg.                                  \\            
                \hline
                \multirow{3}{*}{\begin{tabular}[c]{@{}l@{}}EVA02~\cite{EVA,EVA02}\\ (Large)\end{tabular}} & Full   & 304.24M  & 58.5          & 56.9          & 62.0          & 59.1\cellcolor[HTML]{EFEFEF}          \\
                                                                                                          & Freeze & ~~~0.00M & 54.1          & 51.2          & 54.3          & 53.2\cellcolor[HTML]{EFEFEF}          \\
                                                                                                          & Rein   & ~~~2.53M & \textbf{61.4} & \textbf{58.5} & \textbf{62.0} & \textbf{60.7}\cellcolor[HTML]{EFEFEF} \\
                \hline
                \multirow{3}{*}{\begin{tabular}[c]{@{}l@{}}DINOv2~\cite{Dinov2}\\ (Large)\end{tabular}}   & Full   & 304.20M  & 61.2          & 55.9          & 62.5          & 59.9 \cellcolor[HTML]{EFEFEF}         \\
                                                                                                          & Freeze & ~~~0.00M & 58.9          & 56.4          & 60.3          & 58.5\cellcolor[HTML]{EFEFEF}          \\
                                                                                                          & Rein   & ~~~2.53M & \textbf{63.6} & \textbf{59.0} & \textbf{63.7} & \textbf{62.1}\cellcolor[HTML]{EFEFEF} \\

                \hline
            \end{tabular}%
        }
    }
    \caption{Performance Comparison with the proposed \textbf{Rein with SemFPN~\cite{semfpn}} as Backbones under the \textit{GTAV $\rightarrow$ Cityscapes (Citys) + BDD100K (BDD) + Mapillary (Map)} generalization setting. Models are fine-tuned on GTAV and tested on Cityscapes, BDD100K and Mapillary. The best results are \textbf{highlighted}. $*$ denotes trainable parameters in backbones.}
    \label{tab:vfms_Rein_semfpn}
\end{table}

\begin{table}[htbp]
    \centering
    \renewcommand\arraystretch{1.1}
\setlength\tabcolsep{2pt}
\begin{tabular}{lcc|c}
\hline
Backbone  & Decoder     & Tune & mIoU \\
\hline
ResNet101~\cite{wildnet} & DeeplabV3plus  & Full   & 34.4     \\
ResNet101    & Mask2Former  & Full   &46.1\\
DINOv2    & Mask2Former & Full   & 61.7 \\
DINOv2\cellcolor[HTML]{EFEFEF}    & Mask2Former\cellcolor[HTML]{EFEFEF} & Ours\cellcolor[HTML]{EFEFEF}   & \textbf{64.3}\cellcolor[HTML]{EFEFEF} \\
\hline
\end{tabular}

    \caption{Results on \textit{GTAV$\rightarrow$ Citys+BDD+Map}. Metrics for first line are from Wildnet.}
    \label{tab:benefits}
\end{table}
\section{Multi-source generalization.}
In this part, we compare Rein against other DGSS methods under \textit{GTAV + Synthia $\rightarrow$ Citys + BDD + Map} setting, in which networks are fine-tuned using both GTAV and Synthia datasets, and tested on Cityscapes, BDD100K, and Mapillary. As shown in Table~\ref{tab:gs2cbm}, we report the performance of Rein employing two VFMs, EVA02 and DINOv2. Our results demonstrate that Rein significantly surpasses existing DGSS methods by a large margin in average mIoU (from $45.9\%$ to $65.2\%$).
\setlength{\abovecaptionskip}{0.1cm}
\setlength{\belowcaptionskip}{0.1cm}
\begin{table}[tbp]
    \centering
    \resizebox{\linewidth}{!}{%
        \renewcommand\arraystretch{1.0}
        \setlength\tabcolsep{3.5pt}{
            \begin{tabular}{l|c|cccc}
                \hline
                \multirow{2}{*}{Methods}       & \multirow{2}{*}{Publication} & \multicolumn{4}{c}{mIoU}                                                                                              \\
                \cline{3-6}
                                               &                              & \multicolumn{1}{c}{Citys} & \multicolumn{1}{c}{BDD} & \multicolumn{1}{c}{Map} & \multicolumn{1}{c}{Avg.}              \\
                \hline
                RobustNet~\cite{choi2021robustnet}         & CVPR 21                      & 37.7                      & 34.1                    & 38.5                    & 36.8\cellcolor[HTML]{EFEFEF}          \\
                PintheMem~\cite{PintheMem}      & CVPR 22                      & 44.5                      & 38.1                    & 42.7                    & 41.8\cellcolor[HTML]{EFEFEF}          \\
                SAN-SAW~\cite{SAN-SAW}          & CVPR 22                      & 42.1                      & 37.7                    & 42.9                    & 40.9\cellcolor[HTML]{EFEFEF}          \\
                WildNet~\cite{wildnet}          & CVPR 22                      & 43.7                      & 39.9                    & 43.3                    & 42.3\cellcolor[HTML]{EFEFEF}          \\
                DIGA~\cite{DIGA}                & CVPR 23                      & 46.4                      & 33.9                    & 43.5                    & 41.3\cellcolor[HTML]{EFEFEF}          \\
                SPC~\cite{SPC}                  & CVPR 23                      & 46.4                      & 43.2                    & 48.2                    & 45.9\cellcolor[HTML]{EFEFEF}          \\
                \hline
                EVA02 - Frozen~\cite{EVA,EVA02} & arXiV 23                     & 55.8                      & 55.1                    & 59.1                    & 56.7\cellcolor[HTML]{EFEFEF}          \\
                EVA02 + Rein                   & -                            & 63.5                      & 60.7                    & 63.9                    & 62.7\cellcolor[HTML]{EFEFEF}          \\
                \hline
                DINOv2 - Frozen~\cite{Dinov2}   & arXiV 23                     & 64.8                      & 60.2                    & 65.2                    & 63.4\cellcolor[HTML]{EFEFEF}          \\
                DINOv2 + Rein                  & -                            & \textbf{68.1}             & \textbf{60.5}           & \textbf{67.1}           & \textbf{65.2}\cellcolor[HTML]{EFEFEF} \\
                \hline
            \end{tabular}
        }
    }
    \caption{Performance Comparison of the proposed \textbf{Rein against other DGSS methods} under \textit{ GTAV + Synthia $\rightarrow$ Cityscapes (Citys) + BDD100K (BDD) +Mapillary (Map)} generalization. }
    \label{tab:gs2cbm}
    \vspace{-5mm}
\end{table}
\section{More details about VFMs}
\noindent\textbf{CLIP.} 
In our study, we utilize the ViT-Large architecture, setting the patch size to $16\times16$. Each layer of this architecture outputs features with a dimensionality of 1024, making use of the pre-trained weights from the foundational work~\cite{CLIP}. Our model undergoes a pre-training phase through contrastive learning, employing publicly available image-caption data. This data is compiled through a blend of web crawling from select websites and integrating widely-used, existing image datasets. For the model's pre-trained weights, which have a patch size of $14\times14$ and an original pre-training image size of $224\times224$, we adopt bilinear interpolation to upscale the positional embeddings to a length of 1024. Moreover, trilinear interpolation is utilized to enlarge the kernel size of the patch embed layer to $16\times16$. Our model comprises 24 layers, and the features extracted from the 7th, 11th, 15th, and 23rd layers (counting from the zeroth layer) are subsequently channeled into the decoding head.

\noindent\textbf{MAE.}
Employing the ViT-Large architecture, our model outputs features from each layer with a dimensionality of 1024, maintaining a patch size of $16\times16$. This model capitalizes on the pre-trained weights as delineated in the original work~\cite{MAE}, and it undergoes self-supervised training using masked image modeling on ImageNet-1K. The architecture is composed of 24 layers, directing features from the 7th, 11th, 15th, and 23rd layers directly into the decoding head.

\noindent\textbf{SAM.}
Aligning with the methodology described in the foundational paper~\cite{SAM}, we employ the ViT-Huge architecture as our image encoder, making use of pre-trained weights that were trained on SA-1B~\cite{SAM} for a promptable segmentation task. The patch size of this model is set to $16 \times 16$, and each layer is designed to output features with a dimensionality of 1280, summing up to a total of 32 layers. The positional embeddings of the model are upscaled to a length of 1024 via bicubic interpolation. From this model, we extract features from the 7th, 15th, 23rd, and 31st layers and feed them into the decoder.

\noindent\textbf{EVA02.}
In our approach, we adopt the largest scale configuration, EVA02-L, as our structural backbone, as suggested in the paper~\cite{EVA02}. This particular model configuration determines its patch size as 16, with each layer producing feature maps of 1024 dimensions, across a total of 24 layers. EVA02 undergoes training through a combination of CLIP and Masked Image Modeling techniques on an aggregated dataset that includes IN-21K~\cite{imagenet}, CC12M~\cite{cc12m}, CC3M~\cite{cc3m}, COCO~\cite{coco}, ADE20K~\cite{ade20k}, Object365~\cite{objects365}, and OpenImages~\cite{openimage}. Mirroring the approach used in previous models, we upscale the positional embeddings to 1024 through bilinear interpolation, and the patch embed layer's convolutional kernel size is augmented to $16 \times 16$ via bicubic interpolation. Features from the 7th, 11th, 15th, and 23rd layers are then processed through the decode head.

\noindent\textbf{DINOv2.}
Our choice of backbone for this study is DINOv2-L, which has been distilled from DINOv2-g. As noted in the original documentation, DINOv2-L occasionally surpasses the performance of DINOv2-g~\cite{Dinov2}. Sharing the same patch size, dimensionality, and layer count as EVA02-L, we apply equivalent processing to both the positional embeddings and patch embed layer of DINOv2-L. The features extracted from the 7th, 11th, 15th, and 23rd layers are subsequently fed into the decode head. DINOv2 is originally pretrained in a self-supervised fashion on the LVD-142M~\cite{Dinov2} dataset, following the procedures outlined in its respective paper.

\noindent\textbf{VPT, LoRA, and AdaptFormer.}
Based on extensive experimentation, we have optimized the implementation of PEFT methods for DINOv2, utilizing configurations that enhance performance. These methods include:
1) VPT: It is deep and has 150 tokens.
2) LoRA: Applied to the query and value MLP components, LoRA is configured with a rank of 8. Additionally, it incorporates a minimal dropout rate of 0.1\%.
3) AdaptFormer: This method employs a bottleneck design with a width of 64, initialized using LoRA. Notably, it omits layer normalization.
\section{Algorithm of Proposed Rein}
\label{sec:details_rein}
Algorithm~\ref{Rein} outlines the training procedure for Rein, wherein the weights conform to the constraints specified in Eq.~(11). In this context, the variable $c$ represents the number of channels in the feature maps of model $\mathcal{M}$, $N$ denotes the total number of layers within $\mathcal{M}$, $T$ indicates the overall number of training iterations, and $r$ is defined as a hyperparameter that is considerably smaller than $c$.

\SetCommentSty{mycommfont}
\SetKwInput{KwInput}{Input}
\SetKwInput{KwOutput}{Output}
\begin{algorithm}
\caption{Training process of Rein.}
\label{Rein}
\small
\DontPrintSemicolon
\KwInput{A sequence of input data and corresponding labels $\{(x_i, y_i) \mid t \in \mathbb{N}, 1 \leq i \leq N_d\}$; Pre-trained Vision Foundation Model $\mathcal{M}$, consisting of a patch embed layer $L_{\text{emb}}$, and layers $L_1, L_2, \ldots, L_N$; a decode head $\mathcal{H}$; and a proposed module Rein $\mathcal{R}$. The module Rein comprises the following matrices and vectors, initialized as specified:
\[
\begin{array}{ll}
A_i \in \mathbb{R}^{m \times r}, & \text{uniformly initialized}, \\
B_i \in \mathbb{R}^{r \times c}, & \text{uniformly initialized}, \\
W_{T_i} \in \mathbb{R}^{c \times c}, & \text{uniformly initialized}, \\
W_{f_i} \in \mathbb{R}^{c \times c}, & \text{initialized to zero}, \\
W_{Q_i} \in \mathbb{R}^{c \times c'}, & \text{uniformly initialized}, \\
b_{T_i} \in \mathbb{R}^{c}, & \text{initialized to zero}, \\
b_{f_i} \in \mathbb{R}^{c}, & \text{initialized to zero}, \\
b_{Q_i} \in \mathbb{R}^{c'}, & \text{initialized to zero}, \\
\end{array}
\]
for each $i \in \mathbb{N}, 1 \leq i \leq N$. Additionally, $W_Q \in \mathbb{R}^{3c' \times c'}$ is uniformly initialized, and $b_{Q} \in \mathbb{R}^{c'}$ is initialized to zero.
}
\KwOutput{The optimized $\mathcal{H}$ and $\mathcal{R}$.}

\For{$t\gets1$ \KwTo $T$}{
    Get batch data:$(x,y)$\;
    $f_0=L_\text{emb}(x)$\;
    \For{$i\gets1$ \KwTo $N$}{
        $f_i=L_i(f_{i-1})$\;
        $T_i=A_i\times B_i$\;
        $S_i=Softmax(\frac{f_i\times T_i^\text{T}}{\sqrt{c}})$\;
        $\Delta \Bar{f_i}=S_i(:,2:m)\times[T_i(2:m)\times W_{T_i}+b_{T_i}]$\;
        $\Delta f_i=(\Delta \Bar{f_i}+f_i)\times W_{f_i}+b_{f_i}$\;        
        $Q_i=T_i \times W_{Q_i}+b_{Q_i}$\;
        $f_i=f_i+\Delta f_i$\;
    }
    $\mathbb{F}_t \subseteq \{f_0, f_1, \ldots, f_N\}$\;    
    Calculate $Q_{max}$ and $Q_{avg}$ by Eq.~(9)\;
    $Q=Concat([Q_{max},Q_{avg},Q_N])\times W_Q + b_Q$\;
    $\Bar{y_t}=\mathcal{H}(\mathbb{F}_t,Q)$\;
    Optimize $\mathcal{H}$ and $\mathcal{R}$ by $\mathcal{L}oss(\Bar{y},y)$\;
}
\end{algorithm}
\section{Qualitative Results and Future works}
\begin{figure*}
    \centering
    \includegraphics[width=1.0\linewidth]{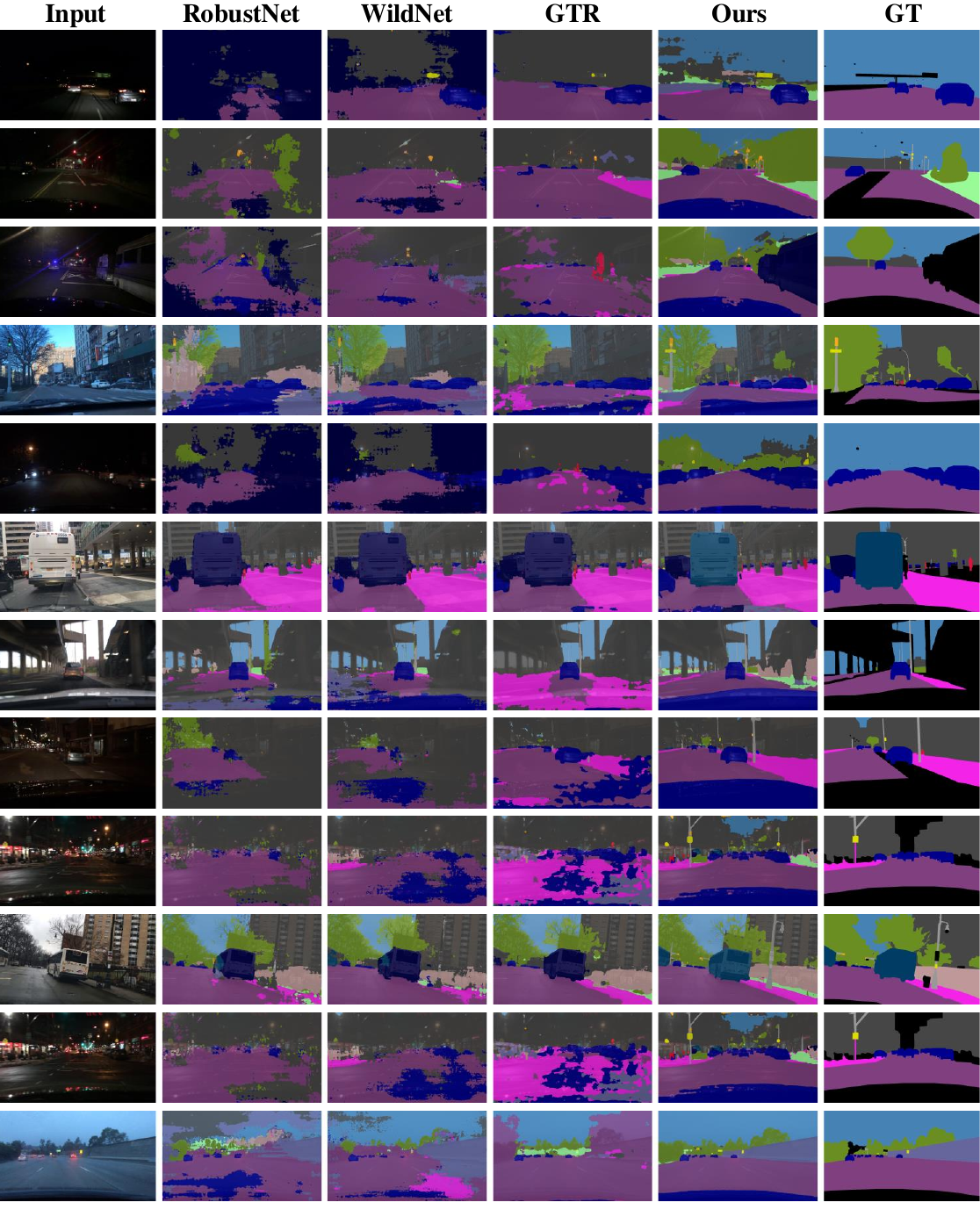}
    \caption{Prediction results of DINOv2+Rein on the BDD100K validation set. The model is fine-tuned exclusively on the GTAV dataset, without access to any real-world urban-scene datasets.}
    \label{fig:bdd}
\end{figure*}

\begin{figure*}
    \centering
    \includegraphics[width=1.0\linewidth]{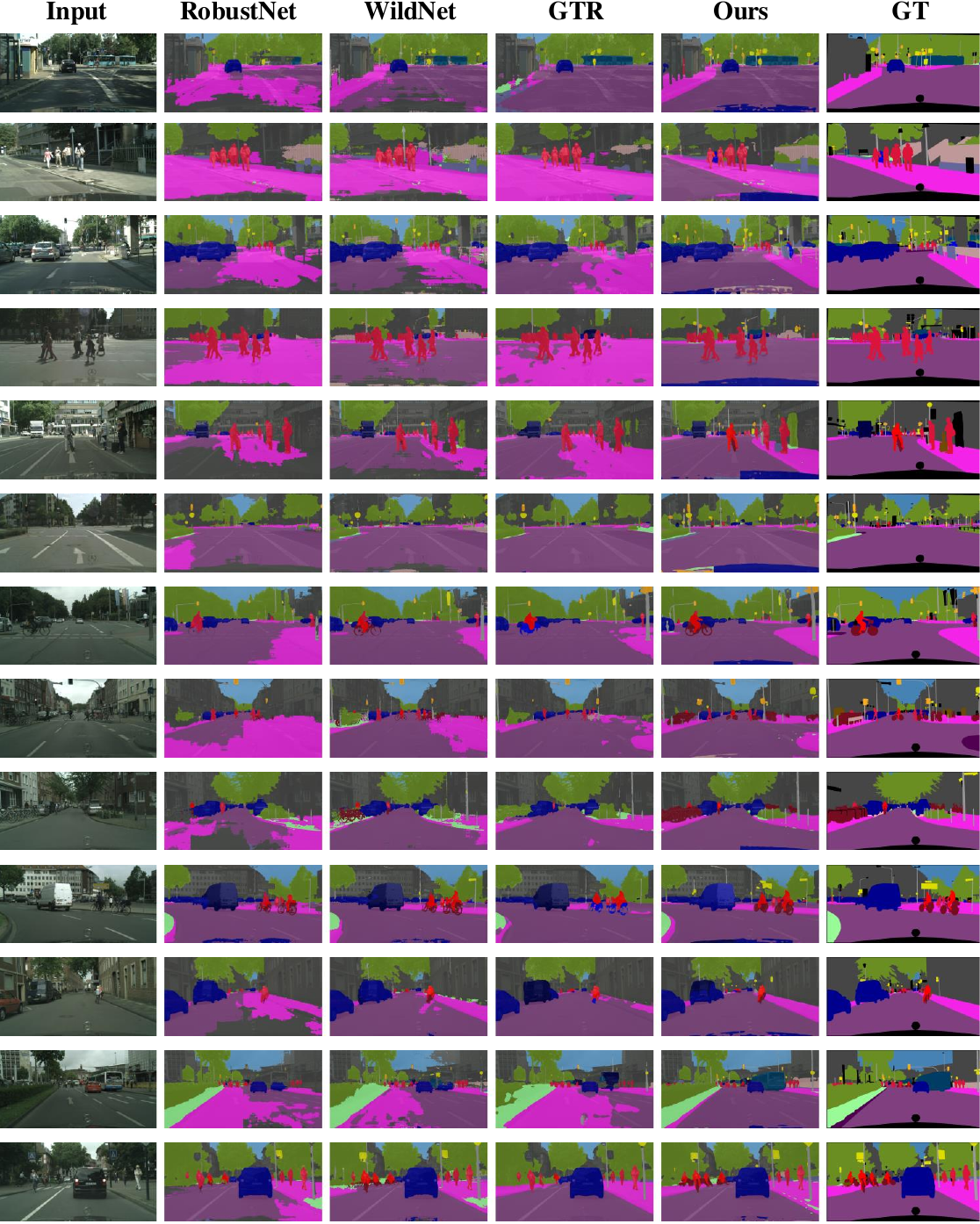}
    \caption{Prediction results of DINOv2+Rein on the Cityscapes validation set. The model is fine-tuned exclusively on the GTAV dataset, without access to any real-world urban-scene datasets.}
    \label{fig:citys}
\end{figure*}

\begin{figure*}
    \centering
    \includegraphics[width=1.0\linewidth]{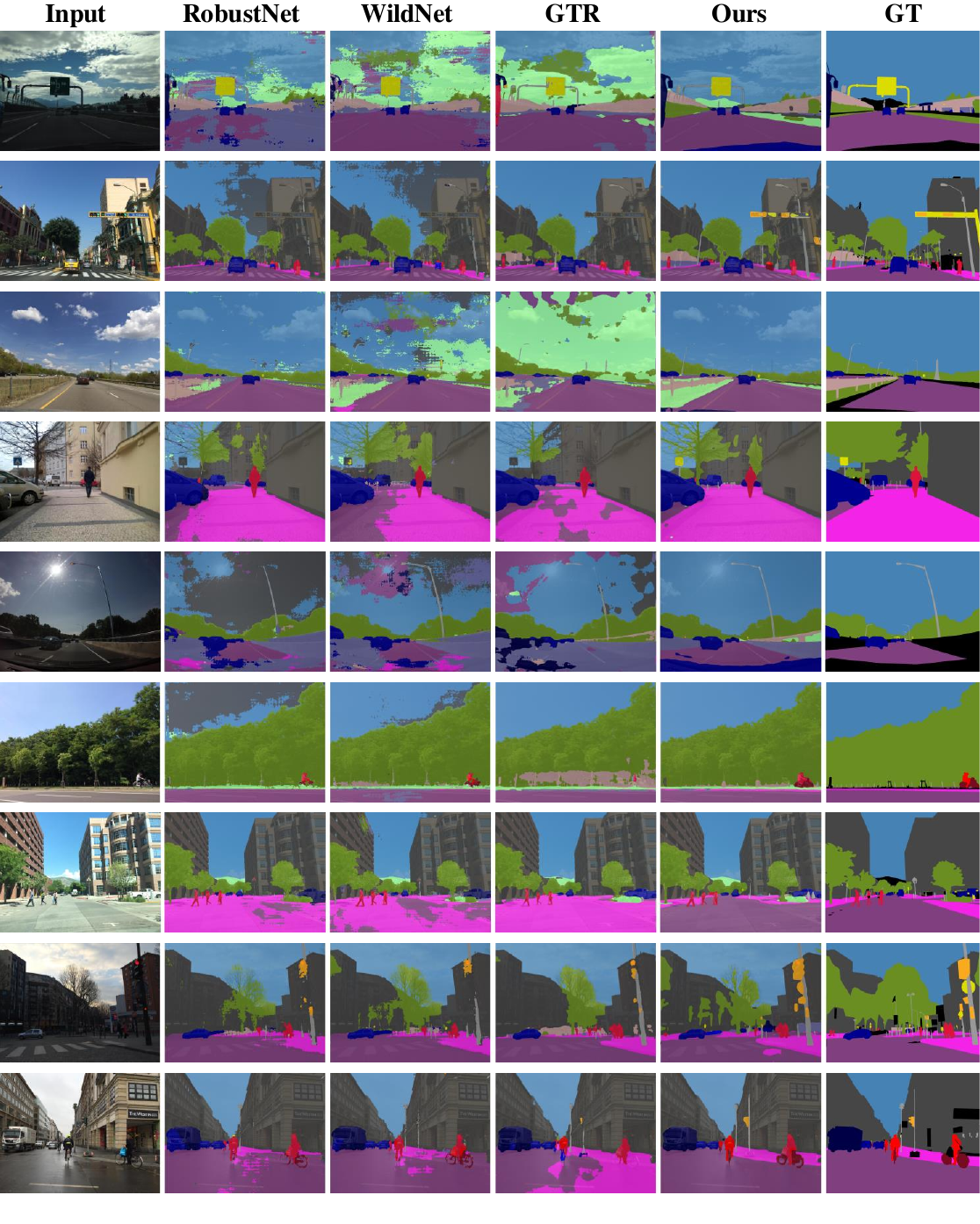}
    \caption{Prediction results of DINOv2+Rein on the Mapillary validation set. The model is fine-tuned exclusively on the GTAV dataset, without access to any real-world urban-scene datasets.}
    \label{fig:map}
\end{figure*}

In this section, we showcase our prediction results across various datasets, including Cityscapes, BDD100K, and Mapillary, as depicted in Fig.\ref{fig:bdd}, Fig.\ref{fig:map}, and Fig.\ref{fig:citys}. All models are trained on the GTAV dataset without any fine-tuning on real-world urban-scene datasets. Our method outshines other approaches in accuracy, especially in categories like traffic signs, bicycles, traffic lights, sidewalks, roads, and trucks, demonstrating high precision for both large objects and smaller targets. Notably, despite not specifically optimizing for night-time segmentation, Rein's performance during night conditions is surprisingly high, almost akin to daytime performance, as illustrated in Fig.\ref{fig:bdd}.

With the rapid development of generative models research, we anticipate that our work could leverage high-quality generated samples to approach the performance of models trained with supervision on real datasets. Furthermore, we are prepared to investigate how VFMs can enhance the performance of semantic segmentation models trained on real datasets under various adverse weather conditions or on special road types. Finally, further exploration is necessary to investigate how Rein can be extended to tasks such as instance segmentation, panoptic segmentation, open-vocabulary segmentation, and even object detection.

\clearpage
{
    \small
    \bibliographystyle{ieeenat_fullname}
    \bibliography{main}
}
\end{document}